\pdfoutput=1

\documentclass[10pt,twocolumn,letterpaper]{article}

\usepackage[pagenumbers]{cvpr} %

\usepackage{multirow}
\usepackage{pifont}                %
\usepackage{colortbl}
\usepackage{bm}

\def\suppmat{Supp.\ Sec.}

\definecolor{baselinebg}{HTML}{F5F5F5} %
\definecolor{oursbg}{HTML}{E1EBF5}     %

\newcommand{\vid}{\mathcal{V}}            %
\newcommand{\bitstream}{\mathcal{B}}      %
\newcommand{\blocks}{\mathcal{M}}         %
\newcommand{\dct}{d}                      %
\newcommand{\finalset}{\mathcal{S}}       %

\newcommand{\score}{s}                    %
\newcommand{\avgscore}{\bar{s}}           %

\newcommand{\grid}{\mathcal{G}}           %
\newcommand{\stride}{\Delta}              %
\newcommand{\thresh}{\theta}              %
\newcommand{\selrate}{R_{\mathrm{sel}}}   %
\newcommand{\vidrate}{R_{\vid}}           %

\newcommand{\methodshort}{CodecSelect}
\newcommand{\methodfull}{\methodshort}

\definecolor{cvprblue}{rgb}{0.21,0.49,0.74}
\usepackage[pagebackref,breaklinks,colorlinks,allcolors=cvprblue]{hyperref}

\def\confName{CVPR}
\def\confYear{2027}

\title{Codec-guided Selection Cheaply Removes Redundant Frames for Video LMMs}

\author{
{Luca Zanella\textsuperscript{1,\dag}
\quad Muhammad Ferjad Naeem\textsuperscript{2,\ddag}
\quad Yongqin Xian\textsuperscript{2}
\quad Elisa Ricci\textsuperscript{1,3}}\\\vspace{-1cm}
\and
{Massimiliano Mancini\textsuperscript{1}
\quad Federico Tombari\textsuperscript{2,4}
\quad Alessio Tonioni\textsuperscript{2}}\\
{\small \textsuperscript{1}University of Trento \quad \textsuperscript{2}Google \quad \textsuperscript{3}Fondazione Bruno Kessler \quad \textsuperscript{4}Technical University of Munich}\\
{\tt\small \url{https://lucazanella.github.io/codecselect/}}\vspace{-0.3cm}
}

\begin{document}
\maketitle
\renewcommand{\thefootnote}{\fnsymbol{footnote}}
\footnotetext[2]{Work done as a Student Researcher at Google. Now at Focoos AI.}
\footnotetext[3]{Now at Prometheus.}
\renewcommand{\thefootnote}{\arabic{footnote}}
\begin{abstract}
For temporal reasoning over videos, Large Multimodal Models (LMMs) commonly rely on dense sampling at a fixed frame rate. Consecutive frames are often nearly identical, yet each sampled frame adds visual tokens that the LMM must encode and attend to. While embedding-based selectors drop these redundant frames, they must first process every candidate frame with a vision backbone, reducing or even undoing the time saved in the LMM. 
In this work, we posit that selection should take place before the LMM or any other model processes the video, \ie, at decoding time. We propose \methodfull{}, a training-free method that exploits, during decoding, two signals already stored in the compressed bitstream: residual energy and intra-block ratio. 
Used as proxies for visual novelty, they mark the changes the encoder already found, so \methodshort{} keeps more frames where the content changes and fewer where it is static. This costs under 1.5\% of the total time per query. Across four LMMs and three temporal reasoning benchmarks, \methodshort{} matches the accuracy of dense sampling while keeping 64\% of the frames and reducing latency per query by 36\%. On two long-video benchmarks, where embedding-based selectors are slower than dense sampling, it answers 14 to 16\% faster with comparable accuracy. 
\end{abstract}

\section{Introduction}
\label{sec:intro}

\begin{figure}[t]
    \centering
    \includegraphics[width=\linewidth]{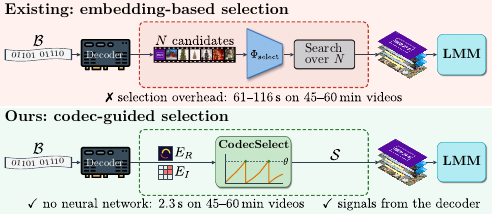}
    \caption{\textbf{Moving frame selection into the video decoder.} \textbf{Top:} embedding-based selectors encode all $N$ candidate frames with a vision backbone $\Phi_{\text{select}}$, then search over them. 
    \textbf{Bottom:} \methodshort{} uses no neural network. It reads two signals the decoder already computes from the bitstream $\bitstream$, $E_R$ and $E_I$, and keeps a frame once they show enough change or enough time has passed.}
    \label{fig:teaser}
    \vspace{-1em}
\end{figure}

Understanding temporal dynamics in videos requires capturing fine-grained motion and state changes over time. To preserve temporal continuity, leading Large Multimodal Models (LMMs), such as Gemini~\cite{comanici2025gemini} and the Qwen series~\cite{bai2025qwen3, qwen35blog}, rely on \textbf{fixed-rate sampling} (\eg, 1 or 2 FPS), which provides denser coverage than traditional sparse strategies~\cite{zohar2025apollo}. However, sampling frames at a constant rate regardless of content introduces substantial redundancy, as consecutive frames often convey near-identical semantics. Every redundant frame still adds visual tokens that the LMM must encode and attend to, so redundancy directly increases memory, compute, and latency.

Adaptive sampling strategies attempt to address this inefficiency. \textbf{Query-dependent selection}~\cite{yu2025frame, buch2025flexible, tang2025adaptive, guo2026logic} filters frames conditioned on the user prompt, reducing tokens per inference pass. However, because the selection process is tied to a specific query, it prevents independent pre-processing, caching, and indexing, requiring the entire video to be re-evaluated for each new prompt. This makes such methods unsuitable for multi-turn interactions and large-scale video retrieval. In contrast, \textbf{query-agnostic selection} aims to identify informative frames independently of the query, allowing for reusable video representations. Yet, existing \textit{embedding-based} methods~\cite{ravi2026spike, li2026maxinfo, song2026ktv, wang2024retake} rely on heuristics that require all candidate frames to be processed through a vision backbone (\eg, CLIP~\cite{radford2021learning}, SigLIP~\cite{zhai2023sigmoid}, or DINOv2~\cite{oquab2024dinov2}) before selection. This introduces a \textbf{selection overhead}, \ie, the cost of encoding every candidate frame with a vision backbone and searching over the result to decide which ones to keep (\cref{fig:teaser}, top). This overhead reduces or even cancels the time saved in the LMM: in our experiments, the slowest selector takes up to $1.8\times$ as long as fixed-rate sampling.

In this work, we propose an alternative design principle: moving \emph{query-agnostic} frame selection into the video decoder. 
We focus on temporal redundancy, as frame-level filtering provides a plug-and-play acceleration mechanism independent of the downstream LMM architecture. To achieve this with minimal overhead, we introduce \textbf{\methodfull{}}, a training-free framework exploiting the spatiotemporal analysis already performed during video compression. At a high level, a video encoder stores each frame as a prediction from other frames plus what this prediction misses, so the bitstream already records where the content changes. \methodshort{} reads these changes instead of the RGB frames, repurposing two signals the video encoder already produces, residual energy and intra-block ratio, as decoder-native proxies for visual novelty (\cref{fig:teaser}, bottom). 
The residual energy measures what the prediction misses, and the intra-block ratio the share of the frame coded without reference to other frames.
Just as video encoders spend bits where the content changes rather than at a constant bitrate, \methodshort{} samples the video at a roughly constant amount of change instead of a constant rate.
Unlike prior adaptive samplers, \methodshort{} requires no neural pre-processing, and its signals are produced by the same decoding pass that every pipeline already runs. This eliminates unnecessary computation at its source and adds only 8\% to the decoding time, under 1.5\% of the total time per query, making it well-suited for low-latency pre-processing and scalable video indexing pipelines. 

We evaluate \methodshort{} across four LMMs (\ie, Qwen3-VL~\cite{bai2025qwen3}, Qwen3.5~\cite{qwen35blog}, InternVL3.5~\cite{wang2025internvl3}, and Gemma~4~\cite{team2026gemma}) on three temporal reasoning benchmarks, TempCompass~\cite{liu2024tempcompass}, MVBench~\cite{li2024mvbench}, and TVBench~\cite{cores2025tvbench}, and on LongVideoBench~\cite{wu2024longvideobench} and Video-MME~\cite{fu2025video}, from short clips to hour-long recordings. Our results show that codec-level selection preserves fixed-rate sampling accuracy while reducing latency, without the selection overhead of embedding-based methods.

\noindent\textbf{Contributions.} To summarize, we make three key {contributions}:
\ding{172} We identify and quantify the \textbf{selection overhead} of existing embedding-based selectors, showing that the cost of processing every candidate frame can negate the savings of the frames they discard.
\ding{173} We propose \textbf{\methodshort{}}, a training-free framework that selects frames during decoding, using decoder-native proxies for visual novelty read from the compressed bitstream. 
\ding{174} We conduct an extensive empirical study across four LMMs and three temporal reasoning benchmarks, demonstrating that \methodshort{} matches the downstream accuracy of fixed-rate sampling while reducing latency per query by an average of 36\%, and by 14\% and 16\% on Video-MME and LongVideoBench, where embedding-based selectors become slower than dense sampling.

\section{Related Work}
\label{sec:related}

\textbf{Frame Selection for Video LMMs.} 
Adaptive selection methods fall into query-dependent and query-agnostic approaches. \textbf{Query-dependent methods}~\cite{yu2025frame, buch2025flexible, tang2025adaptive, guo2026logic, ye2025re, wang2025videotree, hu2025m} condition the selection on the question. Frame-Voyager~\cite{yu2025frame} ranks frame combinations, while AKS~\cite{tang2025adaptive} and T*~\cite{ye2025re} recursively partition the video to locate relevant frames. Each new question requires a new pass over the video, so the selection cannot be cached. \textbf{Query-agnostic methods} select frames from the video alone. SPIKE~\cite{ravi2026spike} allocates more samples to surprising events, MaxInfo~\cite{li2026maxinfo} keeps the frames spanning the largest volume in CLIP space, KTV~\cite{song2026ktv} clusters DINOv2 features into keyframes, and DPSelect, from ReTaKe~\cite{wang2024retake}, keeps the peaks of change in the LMM's own visual tokens. These selections can be precomputed, but they encode every decoded candidate frame with a vision backbone and search over all of them, a \textbf{selection overhead} that grows with video length and reduces or even cancels the time saved in the LMM.

In contrast, \methodshort{} chooses frames during decoding, from signals the decoder already computes, with no neural network and minimal overhead.

\noindent\textbf{Codec Signals for Video Understanding.}
Early methods detected shot boundaries and keyframes from MPEG DCT coefficients and motion vectors without full decoding~\cite{yeo1995rapid, kobla1997compressed}. Later approaches recognized actions directly from motion vectors and residuals~\cite{wu2018compressed, shou2019dmc}. In video LMMs, codec primitives are mainly used in two ways. \textit{Codec-native models} learn to encode them. EMA~\cite{zhao2025efficient} encodes groups of pictures (GOPs) from their I-frames (Sec.~\ref{sec:background}) and motion vectors, Video-LaVIT~\cite{jin2024video} tokenizes keyframes and motion separately, CoPE-VideoLM~\cite{sarkar2026cope}, ReMoRa~\cite{yashima2026remora}, RESUME~\cite{zhang2026resume} and HIMMEL~\cite{jin2026hy} project motion vectors and residuals into the token space through trained adapters, and AdaCodec~\cite{hou2026adacodec} learns tokens for predictable frames. OneVision-Encoder~\cite{tang2026onevision}, Mage-VL~\cite{yang2026mage} and LLaVA-OneVision-2~\cite{an2026llava} select spatiotemporal patches by motion and residual energy, inside models trained on these inputs. \textit{Training-free token reduction} applies codec metadata to off-the-shelf LMMs. MeToM~\cite{wu2026metom} merges low-residual patches and sets a token budget per frame from GOP packet sizes, CodecSight~\cite{zou2026codecsight} prunes patches with motion vectors in streaming VLMs, and Sheng~\etal~\cite{sheng2026video} guide token merging with residuals and motion vectors. Codec information has also been used to choose frames: LoHi~\cite{dong2026rethinking} places its high-resolution frames at the nearest I-frame as one of its variants, and BLUE~\cite{baid2026blue} skips surveillance frames by packet size after re-encoding them with its own codec.

Unlike codec-native models and patch selectors inside trained vision encoders, which change or retrain the LMM, \methodshort{} works with any off-the-shelf LMM. Training-free token reduction prunes or merges patches within frames that are still sampled uniformly and decoded, whereas \methodshort{} removes whole redundant frames, so the two are complementary. Unlike LoHi and BLUE, \methodshort{} is a general selection rule for standard hybrid block-based codecs, such as H.264 and HEVC: it computes the residual energy and intra-block ratio from data their decoders already parse, discards redundant frames before any neural encoding, and requires no training or query.

\section{Preliminaries}
\label{sec:background}

\subsection{Video Coding}
\label{sec:background:coding}

Modern video compression standards (\eg, H.264~\cite{wiegand2003overview}, HEVC~\cite{sullivan2012overview}) achieve high efficiency by exploiting spatial and temporal redundancy via a hybrid block-based approach. Rather than treating frames independently, codecs predict each frame from previously decoded content, within the same frame or from other frames, encoding only the remaining residual and the metadata required to generate those predictions (\eg, prediction modes, block partitions, and motion vectors) into the bitstream. In the following, we review the parts of this process that \methodshort{} reads and to which the rest of the paper refers.

\noindent\textbf{Color Space and Partitioning.} Raw RGB frames are first converted to the YUV color space, separating luminance (Y) from chrominance (U, V). Because the human visual system is less sensitive to color detail than to brightness, chrominance is typically subsampled (\eg, 4:2:0) to reduce data volume. Each frame is partitioned into a grid of coding units (\eg, $16\times16$ macroblocks in H.264), which can be subdivided into smaller blocks to capture fine detail.

\noindent\textbf{Prediction Modes.} Each block uses one of two prediction modes, as allowed by the frame type: 
\begin{enumerate}[label=\textbf{\arabic*)}]
    \item \textbf{Intra-prediction} exploits spatial redundancy by predicting each block from already decoded neighboring blocks within the same frame via texture extrapolation. Frames made only of intra-predicted blocks are called \textbf{I-frames} (Intra-coded). They are decoded without reference to other frames, and those that start a new group of pictures (IDR frames) are random-access points.
    \item \textbf{Inter-prediction} exploits temporal redundancy by referencing previously decoded frames. The encoder searches reference frames for a matching block and records its displacement as a motion vector. \textbf{P-frames} (Predicted) reference previous frames, while \textbf{B-frames} (Bi-predictive) can reference both past and future frames in display order. Importantly, P- and B-frames may still contain intra-predicted blocks wherever spatial prediction is cheaper than the best temporal match, typically where new content appears that no reference contains.
\end{enumerate}

\noindent\textbf{Residuals, Quantization, and Entropy Coding.} The difference between the original pixels and the prediction is the residual. This error signal is transformed into the frequency domain by a block transform (\eg, an integer approximation of the Discrete Cosine Transform), which concentrates signal energy into low-frequency coefficients. These coefficients are then quantized, a lossy step that trades precision for compression. Finally, the quantized coefficients and metadata (\eg, motion vectors) are entropy coded losslessly (\eg, CABAC~\cite{marpe2003context}) into the bitstream.

\noindent\textbf{Decoding Process.} To reconstruct the video, the decoder reverses these steps: entropy decoding, dequantization, and inverse transform, which maps the coefficients back to a pixel-domain residual. This residual is added to the prediction (from inter- or intra-prediction). Finally, loop filters (\eg, deblocking) reduce compression artifacts before display. During these steps, \methodshort{} reads the intra-block ratio (the share of intra-predicted blocks) from the prediction modes and the residual energy from the dequantized coefficients, before the inverse transform (Sec.~\ref{sec:method:proxies}).

\begin{figure*}[t]
    \centering
    \includegraphics[width=\textwidth]{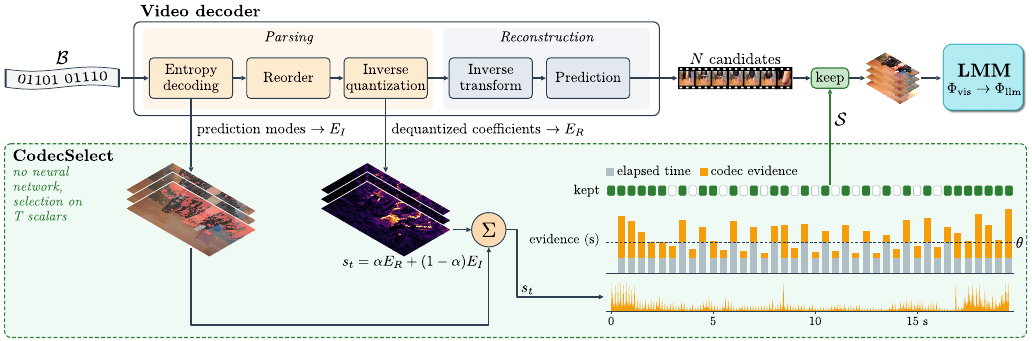}
    \caption{\textbf{Overview of \methodshort{}.} Every pipeline decodes the bitstream $\bitstream$. While it parses, \methodshort{} reads the prediction modes, which give the intra-block ratio $E_I$ (intra-coded blocks in red), and the dequantized coefficients, which give the residual energy $E_R$, and combines them into the novelty score $\score_t$. A grid frame is kept (green) when the evidence since the last kept frame, elapsed time (grey) plus codec evidence (amber), reaches $\thresh$ (here $1$\,s). Only the kept frames $\finalset$ reach the LMM.}
    \label{fig:method_pipeline}
    \vspace{-1em}
\end{figure*}

\subsection{Formalizing the Selection Overhead}
\label{sec:method:formulation}

Let an LMM consist of a vision encoder $\Phi_{\text{vis}}$ and a language model $\Phi_{\text{llm}}$. Videos are stored as compressed bitstreams $\bitstream = \{B_1, \dots, B_T\}$, where $B_t$ is the compressed payload of frame $I_t$. Recovering frames requires parsing the stream (entropy decoding and dequantization) and then reconstructing pixels. Because P- and B-frames are predicted from previously reconstructed references, obtaining frames at a fixed rate generally requires decoding most of the stream. We write $\mathcal{C}_{\text{dec}}(\bitstream)$ for this cost, which every pipeline that feeds RGB frames to the LMM pays. Standard pipelines sample $N \ll T$ candidate frames at a fixed rate (\eg, 1 or 2 FPS) and pass all of them through $\Phi_{\text{vis}}$ at a per-frame cost $\mathcal{C}_{\text{vis}}$. The resulting visual tokens make up most of the input of $\Phi_{\text{llm}}$, whose cost $\mathcal{C}_{\text{llm}}(n)$ grows with the sequence length $n$, quadratically in attention.

To shorten this sequence, adaptive sampling methods first project all $N$ candidate frames into an embedding space with a selection encoder $\Phi_{\text{select}}$, giving $F_t = \Phi_{\text{select}}(I_t)$ for each candidate $t$. This encoder is either the LMM's own vision encoder or a smaller dedicated model. A selection function $g(\cdot)$ then returns the retained indices $\finalset = g(\{F_t\}; Q)$, optionally conditioned on a query $Q$.

As discussed in Sec.~\ref{sec:intro}, conditioning on $Q$ requires re-scanning the video for every interaction, which rules out pre-processing and caching. Because we want video representations to be computed once and reused, we work exclusively in the query-agnostic setting, $\finalset = g(\{F_t\})$.

Bounding the sequence length ($|\finalset| < N$) reduces the language modeling cost to $\mathcal{C}_{\text{llm}}(|\finalset|)$, but for query-agnostic \emph{embedding-based} methods this saving comes at a cost that grows with $N$: every candidate frame must be passed through $\Phi_{\text{select}}$, and $g$ then searches over all $N$ embeddings. With $\mathcal{C}_{\text{select}}$ the per-frame cost of the selection encoder and $\mathcal{C}_{g}(N)$ the cost of $g$ (\eg, $k$-means or a maximum-volume search), which also grows with $N$, total inference costs
\begin{equation}
    \mathcal{O}\big(\mathcal{C}_{\text{dec}}(\bitstream) + \underbrace{N \cdot \mathcal{C}_{\text{select}} + \mathcal{C}_{g}(N)}_{\text{selection}} + |\finalset| \cdot \mathcal{C}_{\text{vis}} + \mathcal{C}_{\text{llm}}(|\finalset|)\big).
    \label{eq:cost_baseline}
\end{equation}
A lightweight $\Phi_{\text{select}}$ (\eg, a small CLIP variant) lowers the per-frame cost $\mathcal{C}_{\text{select}}$ but not the dependence on $N$: every candidate frame is still encoded and searched.%

\section{\methodfull}
\label{sec:method}

\methodshort{} (\cref{fig:method_pipeline}) removes the selection overhead of Eq.~\eqref{eq:cost_baseline} by defining selection on the bitstream itself, $\finalset = g_{\text{codec}}(\bitstream)$, from quantities the decoder already computes while parsing each $B_t$, before reconstruction. Reading these quantities adds a small fraction to the decoding cost (8\% in our measurements, Sec.~\ref{sec:exp:comparison}), and selecting from them costs a few scalar operations per frame, $c_{\text{rule}}$, with no neural computation:
\begin{equation}
    \mathcal{O}\big(\mathcal{C}_{\text{dec}}(\bitstream) + \underbrace{T \cdot c_{\text{rule}}}_{\text{selection}} + |\finalset| \cdot \mathcal{C}_{\text{vis}} + \mathcal{C}_{\text{llm}}(|\finalset|)\big).
    \label{eq:cost_ours}
\end{equation}
Decoding is shared, and selection shrinks from $N$ neural encodings and a search over them to $T$ scalar operations. We first describe the decoder-native proxies that remove the need for a vision backbone (Sec.~\ref{sec:method:proxies}), then the rule that turns them into a frame subset (Sec.~\ref{sec:method:selection}).

\begin{figure*}[t]
    \centering
    \includegraphics[width=\textwidth]{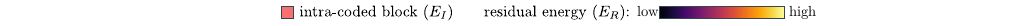} \\
    \vspace{0.1cm}
    \begin{subfigure}[b]{0.32\textwidth}
        \centering
        \captionsetup{justification=centering}
        \includegraphics[width=\textwidth]{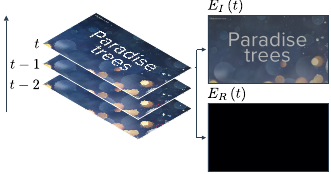}
        \caption{\footnotesize Redundant Frame \\[-0.5ex] $E_R = 0.00$, $E_I = 0.00$}
        \label{fig:novelty_a}
    \end{subfigure}
    \hfill
    \begin{subfigure}[b]{0.32\textwidth}
        \centering
        \captionsetup{justification=centering}
        \includegraphics[width=\textwidth]{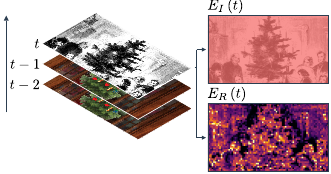}
        \caption{\footnotesize Temporal Discontinuity \\[-0.5ex] $E_R = 1.00$, $E_I = 1.00$}
        \label{fig:novelty_b}
    \end{subfigure}
    \hfill
    \begin{subfigure}[b]{0.32\textwidth}
        \centering
        \captionsetup{justification=centering}
        \includegraphics[width=\textwidth]{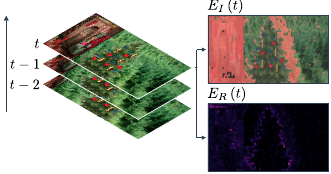}
        \caption{\footnotesize Spatial Novelty \\[-0.5ex] $E_R = 0.02$, $E_I = 0.41$}
        \label{fig:novelty_c}
    \end{subfigure}
    \caption{\textbf{What each signal captures.} Frames $t-2$ to $t$, with the intra-coded blocks ($E_I$) and the residual energy per block ($E_R$) of frame $t$. \textbf{(a)} Predictable motion: almost no intra blocks or residual. \textbf{(b)} A cut starts a new keyframe: all blocks are intra and the residual is the largest of the video. \textbf{(c)} A new scene enters from the left: coded intra ($E_I$ rises), but well predicted within the frame ($E_R$ stays low).
    }
    \label{fig:novelty_proxies}
    \vspace{-1em}
\end{figure*}

\subsection{Decoder-Native Proxies for Visual Novelty}
\label{sec:method:proxies}

To instantiate $g_{\text{codec}}(\bitstream)$ we need proxies for visual novelty that are readable from the bitstream. We obtain them by reusing the analysis the video encoder has already performed. For every block, the bitstream records the prediction mode the encoder chose (Sec.~\ref{sec:background:coding}) and the error left over: together they describe how much of the frame the encoder could not predict from other frames.
\cref{fig:novelty_proxies} shows what each signal captures in three typical cases.

\vspace{2pt}
\noindent\textbf{Residual Energy.} Every block, intra or inter, stores its residual as quantized transform coefficients (Sec.~\ref{sec:background:coding}). Because the block transforms used in modern codecs are approximately orthogonal, and therefore approximately energy-preserving~\cite{malvar2003low}, the magnitude of the prediction error can be measured from the dequantized coefficients, without transforming them back to the pixel domain.

Let $\blocks$ denote the set of coding blocks in $B_t$, and $\dct_{m,u,v}^{(C)}$ the dequantized coefficient of block $m \in \blocks$ at spatial frequency $(u,v)$ in channel $C$. The frame-level residual energy sums squared coefficients over all blocks and channels:
\begin{equation}
    \tilde{E}_R(B_t) = \sum_{C \in \{Y, U, V\}} \sum_{m \in \blocks} \sum_{u,v} \big(\dct_{m,u,v}^{(C)}\big)^2 .
    \label{eq:re_raw}
\end{equation}
The scale of $\tilde{E}_R$ varies with resolution, content, and encoder configuration, so we normalize it over the frame sequence:
\begin{equation}
    E_R(B_t) = \frac{\tilde{E}_R(B_t) - \min_{\tau} \tilde{E}_R(B_\tau)}{\max_{\tau} \tilde{E}_R(B_\tau) - \min_{\tau} \tilde{E}_R(B_\tau)},
    \label{eq:re_norm}
\end{equation}
with $\tau$ ranging over $\{1, \dots, T\}$.

$E_R$ sums over all blocks, intra and inter alike, so it reflects the prediction error left after the encoder has chosen its best mode. It is high for frames that were expensive to code under any prediction: non-linear motion, occlusion, or newly exposed texture that is not self-similar. A low $E_R$, however, is ambiguous: it arises both from genuine redundancy and from new content that happens to be spatially smooth, and therefore cheap to code using intra blocks. 

\vspace{2pt}
\noindent\textbf{Intra-Block Ratio.} We compute $E_I$ as the share of the frame area coded intra. Writing $\blocks_{\text{intra}} \subseteq \blocks$ for the intra-coded subset and $|m|$ for the area of block $m$ in pixels,
\begin{equation}
    E_I(B_t) = \frac{\sum_{m \in \blocks_{\text{intra}}} |m|}{\sum_{m \in \blocks} |m|},
    \label{eq:ibr}
\end{equation}
which lies in $[0,1]$ and reduces to the fraction of intra macroblocks in H.264, where macroblocks have the same size.\footnote{I-frames have $E_I = 1$ by construction and are not treated specially.}

$E_I$ reports the encoder's mode decision rather than the size of its error, \ie, the share of the frame that no reference could predict more cheaply than intra coding. This resolves the ambiguity above: when a new scene enters over a redundant one (\cref{fig:novelty_c}), the references offer no good match for the incoming region, so intra becomes the cheaper mode and $E_I$ rises, while the self-similar incoming texture keeps $E_R$ low. At a hard cut (\cref{fig:novelty_b}) both signals fire.

\vspace{2pt}
\noindent\textbf{Novelty Score.}
The two proxies thus respond to different changes (\cref{fig:novelty_proxies}). Both lie in $[0,1]$, and we combine them as
\begin{equation}
    \score_t = \alpha \, E_R(B_t) + (1-\alpha) \, E_I(B_t), \qquad \alpha \in [0,1].
    \label{eq:novelty}
\end{equation}
The result is a scalar novelty signal $\{\score_t\}_{t=1}^{T}$ over the video, which our selection rule takes as input.

\subsection{Frame Selection}
\label{sec:method:selection}

We use the novelty signal to select a subset of the frames that uniform sampling would give the LMM, so \methodshort{} is a drop-in replacement for fixed-rate sampling. %
Given a target selection rate $\selrate$ in frames per second and a source frame rate $\vidrate$, let $\stride = \vidrate / \selrate$ and $\grid = \{g_1, \dots, g_N\}$, with $g_i = 1 + \lfloor (i-1)\stride \rfloor$, be the uniform grid from the first frame to the end of the video. Consecutive grid frames are one \emph{interval} apart, \ie, $\stride$ frames or $1/\selrate$ seconds. \methodshort{} adds a grid frame to the selection $\finalset \subseteq \grid$ once the evidence that the content has changed since the last kept frame $g_p$ reaches a threshold $\thresh$. This evidence has two terms, the codec evidence and the elapsed time.

\vspace{2pt}
\noindent\textbf{Codec Evidence.} Codec signals are \emph{differential}: $\score_t$ measures how much frame $t$ departs from its reference frames, not from the last kept frame $g_p$. A single $\score_t$ therefore cannot tell whether the content has changed since $g_p$, but summing the novelty of all frames since $g_p$ approximates that change. 
We define the codec evidence at grid frame $g_i$ as $\frac{1}{\avgscore\vidrate}\sum_{j=g_p+1}^{g_i} \score_j$, where $\avgscore = \frac{1}{T}\sum_{t=1}^{T} \score_t$ is the mean novelty per frame, so that it is measured in seconds.

\vspace{2pt}
\noindent\textbf{Elapsed Time.} This approximation misses gradual change. Novelty only compares a frame with its nearby references, so a slow pan or a person turning keeps every $\score_t$, and thus their sum, small while the content drifts away from $g_p$. 
We verify this on the short-video benchmarks of Sec.~\ref{sec:exp}. In one-second windows whose summed novelty is below a quarter of the video's average, the content still changes, in SigLIP~2~\cite{tschannen2025siglip} space, by a median of 43\% of a typical second (\suppmat~\ref{sec:app:selection_strategies}). We therefore add the time elapsed since $g_p$, \ie, $(i-p)/\selrate$ seconds, to the codec evidence. Both are in seconds, so they can be summed and $\thresh$ is a duration.

\noindent\textbf{Selection Rule.} The first grid frame $g_1$ is always kept, and each following one is the first $g_i$, $i > p$, that satisfies:
\begin{equation}
  \underbrace{\frac{1}{\avgscore\vidrate}\sum_{j=g_p+1}^{g_i} \score_j}_{\text{codec evidence}}
  \; + \underbrace{\frac{i - p}{\selrate}}_{\text{elapsed time}} \;\ge\; \thresh .
  \label{eq:trigger}
\end{equation}
The frames kept form $\finalset$, which implements $g_{\text{codec}}(\bitstream)$. Between two kept frames, the evidence is thus about $\thresh$, so \methodshort{} samples the video at a roughly constant amount of change rather than at a constant rate. In a static segment, only time accumulates, so a frame is kept every $\thresh$ seconds, which bounds the gaps. The faster the content changes, the more frames are kept, up to every grid frame.

When a model limits the number of input frames, we raise $\thresh$ for that video until the selection fits. The rule makes two passes over the $T$ scalars, one for $\avgscore$ and the normalization in Eq.~\eqref{eq:re_norm}, and one for selection.

\section{Experiments}
\label{sec:exp}

\begin{table*}[t]
\centering
\small
\setlength{\tabcolsep}{3.2pt}
\caption{\textbf{Comparison with query-agnostic selectors.} Accuracy (\%) on TempCompass (TC), MVBench (MVB), and TVBench (TVB). $^\dagger$: significantly different from dense sampling (McNemar, $p < 0.05$), for Random in all five seeds. Sel.: mean selection time per video. Latency: median time per query relative to dense sampling, averaged over benchmarks and LMMs. }
\label{tab:main}
\resizebox{\linewidth}{!}{
\begin{tabular}{@{}l ccc ccc ccc ccc cc@{}}
\toprule
 & \multicolumn{3}{c}{\textbf{Qwen3-VL}} & \multicolumn{3}{c}{\textbf{Qwen3.5}} & \multicolumn{3}{c}{\textbf{InternVL3.5}} & \multicolumn{3}{c}{\textbf{Gemma~4}} & \multirow{2}{*}{\textbf{Sel. (s)}} & \multirow{2}{*}{\textbf{Latency}} \\
\cmidrule(lr){2-4} \cmidrule(lr){5-7} \cmidrule(lr){8-10} \cmidrule(lr){11-13}
\textbf{Method} & TC & MVB & TVB & TC & MVB & TVB & TC & MVB & TVB & TC & MVB & TVB & & \\
\midrule
\rowcolor{baselinebg} Dense (Uniform 2 FPS) & 75.9 & 67.1 & 59.8 & 73.5 & 67.7 & 61.5 & 71.7 & 72.9 & 62.1 & 73.1 & 56.9 & 49.5 & 0 & 1.00$\times$ \\
\addlinespace[2pt]
\rowcolor{oursbg} \textbf{\methodshort{} (ours)} & 75.9 & 66.5 & 59.3 & 74.0 & 67.0 & 60.8 & 71.2 & 72.6 & 61.6 & 74.0$^\dagger$ & 57.8 & 51.0$^\dagger$ & 0.03 & 0.64$\times$ \\
\addlinespace[4pt]
\multicolumn{15}{@{}l}{\textit{Embedding-based selectors, same number of frames per video as \methodshort{}}} \\
 MaxInfo~\cite{li2026maxinfo} & 75.5 & 66.4 & 58.4$^\dagger$ & 74.0 & 67.2 & 60.0$^\dagger$ & 71.3 & 72.4 & 60.8$^\dagger$ & 74.2$^\dagger$ & 57.2 & 51.8$^\dagger$ & 0.12 & 0.71$\times$ \\
 KTV~\cite{song2026ktv} & 76.0 & 66.7 & 59.0 & 74.3$^\dagger$ & 66.8 & 60.6 & 71.8 & 71.9$^\dagger$ & 61.8 & 74.0$^\dagger$ & 58.1$^\dagger$ & 51.4$^\dagger$ & 1.24 & 1.30$\times$ \\
 DPSelect~\cite{wang2024retake} & 75.4$^\dagger$ & 66.1$^\dagger$ & 59.4 & 73.9 & 67.0 & 59.6$^\dagger$ & 71.3 & 71.4$^\dagger$ & 59.8$^\dagger$ & 74.1$^\dagger$ & 56.7 & 49.2 & 0.29 & 0.75$\times$ \\
\addlinespace[4pt]
\multicolumn{15}{@{}l}{\textit{Fast selectors, without a neural network, at their own setting}} \\
 PySceneDetect~\cite{castellano2026pyscenedetect} & 64.2$^\dagger$ & 51.0$^\dagger$ & 41.4$^\dagger$ & 64.5$^\dagger$ & 52.0$^\dagger$ & 40.5$^\dagger$ & 63.2$^\dagger$ & 52.3$^\dagger$ & 41.6$^\dagger$ & 62.7$^\dagger$ & 48.2$^\dagger$ & 38.9$^\dagger$ & 0.29 & 0.24$\times$ \\
 I-frames~\cite{dong2026rethinking} & 65.1$^\dagger$ & 49.6$^\dagger$ & 42.9$^\dagger$ & 64.8$^\dagger$ & 52.8$^\dagger$ & 45.3$^\dagger$ & 62.5$^\dagger$ & 52.1$^\dagger$ & 45.0$^\dagger$ & 63.7$^\dagger$ & 47.7$^\dagger$ & 42.5$^\dagger$ & 0.02 & 0.15$\times$ \\
\addlinespace[4pt]
\multicolumn{15}{@{}l}{\textit{Controls, same number of frames as \methodshort{}, per video or in total}} \\
 Random (5 seeds) & 75.5 & 65.5$^\dagger$ & 58.2 & 73.8 & 66.5 & 59.9 & 70.8 & 71.3$^\dagger$ & 61.0 & 73.5 & 56.8 & 51.0 & -- & -- \\
 Evenly spaced & 75.8 & 66.5 & 58.9 & 73.6 & 67.0 & 60.6 & 71.3 & 72.4 & 62.2 & 74.2$^\dagger$ & 57.3 & 52.2$^\dagger$ & -- & -- \\
 Uniform 1.28\,FPS, same total frames & 75.8 & 66.5 & 59.0 & 73.5 & 67.2 & 60.5 & 71.2 & 72.4 & 61.5 & 74.1$^\dagger$ & 57.5 & 51.2$^\dagger$ & -- & -- \\
\bottomrule
\end{tabular}}
\end{table*}

We describe the setup, compare \methodshort{} with dense sampling and existing selectors (Sec.~\ref{sec:exp:comparison}), and analyze its design (Sec.~\ref{sec:exp:ablation}).

\begin{figure}[t]
    \centering
    \includegraphics[width=\linewidth]{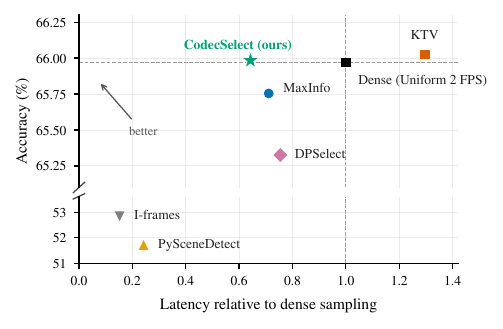}
    \caption{\textbf{Accuracy vs latency}, averaged over the LMMs and benchmarks of \cref{tab:main}. Latency relative to dense sampling. }
    \label{fig:main_tradeoff}
    \vspace{-1em}
\end{figure}

\noindent\textbf{Setup.}
We evaluate four LMMs, Qwen3-VL-8B-Instruct~\cite{bai2025qwen3}, Qwen3.5-9B~\cite{qwen35blog}, InternVL3.5-8B~\cite{wang2025internvl3}, and Gemma-4-12B~\cite{team2026gemma}, on three temporal reasoning benchmarks, where answers depend on how the video unfolds: \textit{TempCompass}~\cite{liu2024tempcompass} (multiple-choice, yes/no, and caption matching), \textit{MVBench}~\cite{li2024mvbench}, and \textit{TVBench}~\cite{cores2025tvbench}. For long videos, we use \textit{Video-MME}~\cite{fu2025video}, without one video that a baseline fails to decode, and the 1,337 validation questions of \textit{LongVideoBench}~\cite{wu2024longvideobench}, from short clips to hour-long recordings, where every method receives at most 960 frames. All videos are re-encoded to H.264 (CRF 23, 3 B-frames), and a patched FFmpeg decoder exports $E_R$ and $E_I$. 
Dense sampling and \methodshort{} use a 2 FPS grid, the default rate of the Qwen-VL series, with $\alpha = 0.5$ and $\thresh = 1$\,s for every LMM and benchmark. Within the frame limit, \methodshort{} never samples below one frame per second, and keeps a frame half a second after the previous one only when the content changed more than in an average half-second of the video. All methods give the LMM their frames as separate timestamped images (\suppmat~\ref{sec:app:implementation}).

\noindent\textbf{Metrics.}
We report accuracy with each benchmark's official protocol and compare two methods on the same questions with McNemar's test. \textit{Latency} is the time per query from the frames to the first answer token, \ie, frame selection plus the LMM (vision encoder and prefill), measured on NVIDIA A100 (64\,GB) GPUs. Selection runs once per video and is shared by its questions. Decoding is required by every method and not counted, except for the signal extraction of \methodshort{} (\suppmat~\ref{sec:app:metrics}).

\noindent\textbf{Baselines.}
\textbf{Dense sampling} keeps every frame of the 2 FPS grid and is the accuracy we aim to preserve. \textit{Embedding-based selectors} encode the decoded candidate frames with a neural network and select from the resulting embeddings: \textbf{MaxInfo}~\cite{li2026maxinfo} keeps the frames spanning the largest volume in CLIP space, the keyframe stage of \textbf{KTV}~\cite{song2026ktv} clusters DINOv2 features, and \textbf{DPSelect}~\cite{wang2024retake} keeps the peaks of change in the LMM's own visual tokens, which the LMM then reuses, as in its official code. It still encodes every candidate frame to choose among them, so the reuse only saves the encoding of the kept frames. To isolate selection in cost and accuracy, %
each of them keeps, for every video, as many frames as \methodshort{}. \textit{Fast selectors} use no neural network and run at their own setting: \textbf{PySceneDetect}~\cite{castellano2026pyscenedetect} keeps one frame per detected shot, and \textbf{I-frames} keeps the video encoder's keyframes, as the anchor variant of LoHi~\cite{dong2026rethinking}. All baselines follow their official code, and embedding-based ones run at best speed (\suppmat~\ref{sec:app:baselines}).

\subsection{Comparison with Query-Agnostic Selectors}
\label{sec:exp:comparison}

\begin{figure}[t]
    \centering
    \includegraphics[width=\linewidth]{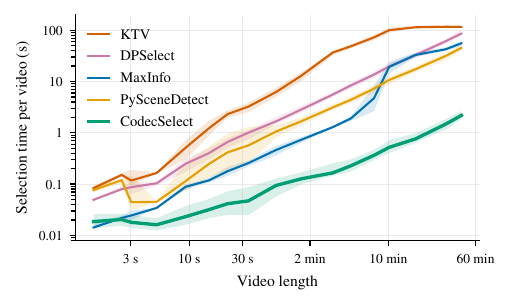}
    \caption{\textbf{Selection cost grows with video length.} Median selection time per video and interquartile band (8,230 videos). \methodshort{} includes the signal export. }
    \label{fig:scaling}
    \vspace{-1em}
\end{figure}

\noindent\textbf{\methodshort{} preserves dense accuracy.}
We compare the methods with the four LMMs on the three benchmarks in \cref{tab:main,fig:main_tradeoff}. \methodshort{} keeps 64\% of the frames on average and matches the accuracy of dense sampling, 66.0 for both when averaged over LMMs and benchmarks. It is never more than 0.7 points below dense sampling, and none of the twelve pairs of LMM and benchmark shows a significant drop. Over all questions and LMMs, its difference from dense sampling is $+0.05$ points (95\% confidence interval $-0.18$ to $+0.29$). Gemma~4 even gains significantly on TempCompass and TVBench. On a 4\,FPS grid with the same $\thresh$, \methodshort{} keeps 38\% of the frames and matches dense 4\,FPS sampling on average (67.9 against 68.0, three LMMs), with one significant drop in nine pairs.

\vspace{2pt}
\noindent\textbf{Embedding-based selectors have high overhead.} %
At the same number of frames, MaxInfo and KTV are about as accurate as \methodshort{}, within 0.3 points of it on average, and DPSelect is 0.7 points below (\cref{tab:main}). However, they differ in the selection overhead. On these short clips, \methodshort{} selects in 0.03\,s per video and the embedding-based selectors in 0.12 to 1.24\,s, because they encode every candidate frame and then search over all of them. The time of \methodshort{} includes exporting the codec signals, which adds a median of 8\% to decoding, under 1.5\% of the total time per query with any of the four LMMs (\suppmat~\ref{sec:app:efficiency_breakdown}). Per query, \methodshort{} answers in 0.64$\times$ the time of dense sampling and MaxInfo in 0.71$\times$, DPSelect in 0.75$\times$, and KTV in 1.30$\times$, slower than not selecting at all (\cref{tab:main}). %

\vspace{2pt}
\noindent\textbf{Fast selectors lose key content.}
PySceneDetect and I-frames answer faster than \methodshort{} only because they keep far fewer frames. On these benchmarks most clips contain a single shot and few keyframes, so they keep one to three frames and lose 13 to 14 points on average, 10 to 18 per benchmark (\cref{tab:main}). Among the selectors we evaluate, \methodshort{} is the fastest that preserves dense accuracy. MaxInfo comes closest, at a higher latency and with a significant drop on TVBench for three of the four LMMs.

\noindent\textbf{Controls at the same number of frames.}
We compare \methodshort{} with two controls that keep, for every video, as many frames as \methodshort{} from the same grid (\cref{tab:main}): \textbf{Random} frames (five seeds) and \textbf{Evenly spaced} frames. \methodshort{} is more accurate than random frames in all twelve pairs, by 0.66 points on average (one-sided paired $t$-test, $p < 0.001$), and matches evenly spaced frames ($-0.01$ points, with no significant difference in any pair). Random frames can leave long stretches of the video uncovered, whereas \methodshort{} never leaves a gap longer than $\thresh$. Evenly spaced frames avoid such gaps too, but need the number of frames per video in advance, while \methodshort{} derives it from the codec signals at almost no cost. 
A fixed lower rate must also be chosen in advance. At 1.28\,FPS, which matches the total number of frames of \methodshort{}, uniform sampling is as accurate in all twelve pairs (\cref{tab:main}). On LongVideoBench, however, it is significantly less accurate (65.7 vs. 67.4, McNemar $p = 0.002$, \cref{tab:videomme}).

\begin{table}[t]
\centering
\caption{\textbf{Comparison on long videos} with Qwen3-VL-8B. Fr.: frames per question. Sel.: mean selection time per video (s). Lat.: mean latency per query (s). $^\dagger$: significantly different from dense sampling (McNemar, $p < 0.05$).}
\label{tab:videomme}
\resizebox{\linewidth}{!}{
\begin{tabular}{@{}l cccc cccc@{}}
\toprule
 & \multicolumn{4}{c}{\textbf{Video-MME}} & \multicolumn{4}{c}{\textbf{LongVideoBench}} \\
\cmidrule(lr){2-5} \cmidrule(lr){6-9}
\textbf{Method} & Acc. & Fr. & Sel. & Lat. & Acc. & Fr. & Sel. & Lat. \\
\midrule
\rowcolor{baselinebg} Dense (Uniform 2 FPS) & 71.3 & 654 & 0 & 72 & 65.9 & 647 & 0 & 75 \\
\rowcolor{oursbg} \textbf{\methodshort{} (ours)} & 71.2 & 569 & 0.9 & 62 & 67.4$^\dagger$ & 563 & 0.5 & 63 \\
 Uniform, same total frames & 71.0 & 570 & -- & -- & 65.7 & 563 & -- & -- \\
 MaxInfo & 68.4$^\dagger$ & 569 & 21.6 & 83 & 63.3$^\dagger$ & 563 & 11.3 & 80 \\
 KTV & 71.0 & 569 & 71.6 & 133 & 65.5 & 563 & 48.3 & 132 \\
 DPSelect & 70.3$^\dagger$ & 569 & 33.0 & 88 & 66.0 & 563 & 14.9 & 79 \\
 PySceneDetect & 67.0$^\dagger$ & 196 & 16.7 & 33 & 61.8$^\dagger$ & 114 & 8.0 & 19 \\
 I-frames & 67.2$^\dagger$ & 228 & 1.2 & 20 & 62.4$^\dagger$ & 142 & 0.1 & 8 \\
\bottomrule
\end{tabular}}
\end{table}

\noindent\textbf{Selection cost matters most on long videos.}
The difference in selection cost grows with video length (\cref{fig:scaling}), since longer videos have more candidates. On videos of 45 to 60 minutes, the embedding-based selectors need a median of 61 to 116\,s per video and \methodshort{} 2.3\,s. On these benchmarks (\cref{tab:videomme}), we report the mean latency per query, since their questions are on either short or very long videos and the median falls in the gap between the two (\suppmat~\ref{sec:app:metrics}). On Video-MME, \methodshort{} matches dense sampling (71.2 against 71.3) and answers in 62\,s instead of 72\,s, although the 960-frame cap already shortens dense sampling. On LongVideoBench, it is more accurate than dense sampling (67.4 against 65.9, McNemar $p = 0.013$) and than uniform sampling at 1.24\,FPS, with the same total number of frames (65.7, $p = 0.002$), and answers in 63\,s instead of 75\,s. Evenly spaced frames at its per-video count lie in between (66.5, \suppmat~\ref{sec:app:tables}). The embedding-based selectors spend 11 to 72\,s per video on selection, so they are slower than dense sampling. PySceneDetect and I-frames keep far fewer frames, making them faster but less accurate.

\subsection{Ablation Studies}
\label{sec:exp:ablation}

\begin{figure}[t]
    \centering
    \includegraphics[width=\linewidth]{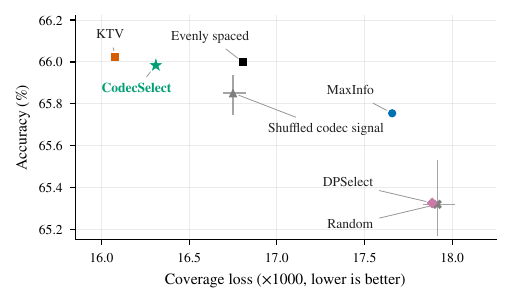}
    \caption{\textbf{Coverage predicts accuracy.} Coverage loss (no LMM) against accuracy (mean over LMMs and benchmarks), all at the number of frames of \methodshort{}. Bars: range over seeds. }
    \label{fig:coverage}
    \vspace{-1em}
\end{figure}

We run the ablations on the three short benchmarks.

\noindent\textbf{Coverage and accuracy.}
At the same number of frames, the benchmarks separate selections only by a few tenths of a point. We therefore also measure, without any LMM, how well the kept frames represent the video, as the cosine distance in SigLIP~2~\cite{tschannen2025siglip} space between each frame and the nearest kept frame in time, averaged per benchmark, then over the benchmarks. We add a third control, the same rule on the codec signal shuffled in time (three seeds). The methods that cover the video better are more accurate (Spearman 0.89, \cref{fig:coverage}). Random frames, MaxInfo, and DPSelect cover the video worst and are the least accurate, since random, diverse, or peak frames can leave stretches of the video far from any kept frame. KTV covers it best, partly because it picks among all frames. \methodshort{} has a 3.0\%, 9.0\%, and 2.6\% lower coverage loss than evenly spaced, random, and shuffled frames, but \methodshort{}, KTV, evenly spaced, and shuffled frames are within 0.2 points of each other in accuracy, too close for the benchmarks to tell apart.

\begin{table}[t]
\centering
\caption{\textbf{Building up \methodshort{}.} Accuracy (\%) over the four LMMs, at the number of frames of \methodshort{}. Kept: frames relative to dense sampling. Coverage: loss of \cref{fig:coverage} ($\times 1000$).}
\label{tab:buildup_count}
\resizebox{\linewidth}{!}{
\begin{tabular}{@{}ccc ccc ccc@{}}
\toprule
\textbf{$E_I$} & \textbf{$E_R$} & \textbf{Time} & \textbf{TC} & \textbf{MVB} & \textbf{TVB} & \textbf{Avg} & \textbf{Kept} & \textbf{Coverage} \\
\midrule
\ding{51} & & & 73.5 & 64.7 & 57.0 & 65.05 & 62\% & 18.14 \\
& \ding{51} & & 73.2 & 65.3 & 57.8 & 65.45 & 65\% & 18.68 \\
\ding{51} & \ding{51} & & 73.3 & 65.5 & 57.6 & 65.46 & 64\% & 17.57 \\
\rowcolor{oursbg} \ding{51} & \ding{51} & \ding{51} & 73.8 & 66.0 & 58.2 & 65.98 & 64\% & 16.31 \\
\bottomrule
\end{tabular}}
\end{table}

\noindent\textbf{Component analysis.} 
We add the components of \methodshort{} one at a time, at about the same number of frames (62 to 65\% of the grid), so that mainly frame placement changes (\cref{tab:buildup_count}). %
$E_I$ alone is the least accurate, and adding $E_R$ to it raises accuracy by 0.4 points. $E_R$ alone is about as accurate as the combination (65.45 against 65.46), but it covers the video worse: the combination has a 5.9\% lower coverage loss than $E_R$ alone and 3.1\% lower than $E_I$ alone. The two signals miss different events, \eg, $E_R$ barely responds when a scene fades in, and $E_I$ barely responds to a change that the video encoder predicts with a large residual (\suppmat~\ref{sec:app:visualizations}). Adding the elapsed time raises accuracy by a further 0.5 points and lowers the coverage loss by 7.2\%, since it keeps frames in the segments of gradual change that the codec signals miss. Keeping instead the grid frames with the highest novelty, at the same count, is 0.44 points less accurate (65.55 against 65.98, lower in 10 of 12 pairs, $p = 0.004$) and covers the video 16\% worse, leaving gaps over 2\,s in a third of the videos.
We use $\alpha = 0.5$, and other values give the same accuracy (\suppmat~\ref{sec:app:tables}).

\noindent\textbf{Robustness to the encoding.}
We re-encode the three benchmarks with HEVC instead of H.264, without B-frames, and at a higher (CRF~18) and a much lower (CRF~35) quality, and compare \methodshort{} with dense sampling on the same videos (\cref{tab:encoding}). Unlike H.264 macroblocks, HEVC blocks vary in size, but $E_I$ already weighs each block by its area (Sec.~\ref{sec:method:proxies}), so the method needs no change. Across encodings, \methodshort{} keeps 62 to 64\% of the frames and stays within 0.4 to 0.8 points of dense sampling on average. It also selects largely the same frames: under each other encoding, it keeps 87 to 96\% of the frames it keeps under the default H.264 encoding, 
against 67\% for the shuffled codec signal and 62\% for random frames. 
\begin{table}[t]
\centering
\caption{\textbf{Robustness to the encoding.} Qwen3-VL-8B, accuracy averaged over the three benchmarks. $\Delta$: \methodshort{} minus dense sampling on the same videos. Kept: frames kept by \methodshort{}.}
\label{tab:encoding}
\resizebox{\linewidth}{!}{
\begin{tabular}{@{}llc cc cc@{}}
\toprule
\textbf{Codec} & \textbf{B-frames} & \textbf{CRF} & \textbf{Dense} & \textbf{\methodshort{}} & $\boldsymbol{\Delta}$ & \textbf{Kept} \\
\midrule
\rowcolor{oursbg} H.264 & \ding{51} & 23 & 67.6 & 67.2 & $-0.4$ & 64\% \\
\addlinespace[3pt]
HEVC & \ding{51} & 23 & 67.8 & 67.1 & $-0.7$ & 62\% \\
\addlinespace[3pt]
H.264 & \ding{55} & 23 & 67.6 & 67.1 & $-0.5$ & 64\% \\
\addlinespace[3pt]
H.264 & \ding{51} & 18 & 67.7 & 66.9 & $-0.8$ & 64\% \\
H.264 & \ding{51} & 35 & 67.9 & 67.4 & $-0.5$ & 62\% \\
\bottomrule
\end{tabular}}
\end{table}

\section{Conclusion}
\label{sec:conclusion}

In this work, we formalize the \textit{selection overhead} of query-agnostic frame selection, the cost of encoding every decoded candidate frame and searching over the result, and introduce \methodfull{}, a training-free method that reads two signals the video encoder stores in the bitstream and keeps more frames where the content changes, for under 1.5\% of the total time per query. Across four LMMs and three temporal reasoning benchmarks, \methodshort{} matches the accuracy of dense sampling while keeping 64\% of the frames and cutting latency per query by 36\% on average. On long videos, where embedding-based selectors become slower than dense sampling, it answers 14 to 16\% faster without a significant loss of accuracy. It works with off-the-shelf LMMs and complements codec-guided token reduction.

\noindent\textbf{Acknowledgments.} We acknowledge ISCRA for awarding this project access to the LEONARDO supercomputer, owned by the EuroHPC Joint Undertaking, hosted by CINECA (Italy).

{
    \small
    \bibliographystyle{ieeenat_fullname}
    \bibliography{main}
}

\twocolumn[\centering\Large\textbf{Supplementary Material}\vspace{1em}]
\setcounter{section}{0}
\renewcommand{\thesection}{\Alph{section}}
\makeatletter\setlength{\@fptop}{0pt}\setlength{\@dblfptop}{0pt}\makeatother

In this supplementary material, we first give more details on the implementation (\cref{sec:app:implementation}), benchmarks (\cref{sec:app:datasets}), evaluation protocol (\cref{sec:app:metrics}), and baselines (\cref{sec:app:baselines}) of our experiments. We then analyze the gradual change that the codec signals miss and that the elapsed time covers (\cref{sec:app:selection_strategies}), break down the cost of \methodshort{} (\cref{sec:app:efficiency_breakdown}), and report detailed results, per LMM, at other frame rates, on long videos, and for other values of $\alpha$ and $\thresh$ (\cref{sec:app:tables}). Finally, we show what each codec signal captures and which frames \methodshort{} keeps on real videos (\cref{sec:app:visualizations}), discuss the limitations (\cref{sec:app:limitations}) and broader impact (\cref{sec:app:broader_impact}) of our work, and list the licenses of all assets we use (\cref{sec:app:assets}).

\section{Implementation Details}
\label{sec:app:implementation}

\noindent\textbf{Codec signals.}
We patch the H.264 and HEVC decoders of FFmpeg 9.0.2 so that, while they decode a frame, they also write out the residual energy of each color channel and the area of the intra-coded blocks. Both come from data the decoder parses anyway, so no second pass over the video is needed, and the export adds a median of 8\% to the decoding time (\cref{sec:app:efficiency_breakdown}). In H.264, all macroblocks have the same area, so $E_I$ is the fraction of intra-coded macroblocks. In HEVC, coding units differ in size, and $E_I$ weighs each by its area. 

\noindent\textbf{Encoding.}
The benchmarks distribute their videos in different containers and with different encoder settings, \eg, part of TVBench~\cite{cores2025tvbench} comes as AVI files. We therefore re-encode every video with the same settings, which gives every method the same input and keeps the codec signals comparable across benchmarks. We use x264 at its default preset (H.264 Main profile, CRF 23, up to 3 B-frames, and closed GOPs) and remove the audio. Keyframes are placed by x264's default rules. The HEVC streams of \cref{sec:exp:ablation} are encoded with x265 at the same CRF and B-frames, and the other variants change one setting at a time (no B-frames, CRF 18, or CRF 35). \methodshort{} stays close to dense sampling under these encodings (\cref{tab:encoding}) and on the original videos of Video-MME's temporal questions (\cref{sec:app:long_videos}).

\noindent\textbf{Frame input.}
Every method, dense sampling included, gives its frames to the LMM in the same way. We decode the selected frames with Decord at the resolution of the video, store them as JPEG files as VLMEvalKit\footnote{\url{https://github.com/open-compass/VLMEvalKit}} does, and pass them as separate images in temporal order, each preceded by its timestamp as text, \eg, \texttt{[00:03.500]}. The kept frames are not evenly spaced, so the timestamps tell the LMM when each frame was taken. We use the image input of the LMMs rather than their video input, which assumes evenly spaced frames and can merge or resample them, and would thus change which frames reach the LMM. The prompts are those of VLMEvalKit for each benchmark.

\noindent\textbf{LMMs.}
We run all LMMs through VLMEvalKit in bfloat16 with greedy decoding. InternVL3.5 encodes each frame as a single $448 \times 448$ tile (256 visual tokens), the setting of VLMEvalKit for videos, and Gemma~4 uses the default of its image processor, at most 280 visual tokens per frame. By default, Qwen3-VL and Qwen3.5 encode each frame at its native resolution. We limit them to 256 visual tokens per frame, in line with the other two LMMs, so that the number of tokens grows with the number of frames rather than with the resolution of the video. Qwen3.5 thinks before answering by default, and we disable this thinking mode through its chat template, as it is disabled by default in Gemma~4. For the latency of Gemma~4, every method, dense sampling included, runs on two GPUs with the layers split between them, since its longest inputs do not fit on one GPU.

\noindent\textbf{Frame cap on long videos.}
On Video-MME and LongVideoBench, every method receives at most 960 frames per video. With this cap, the longest input to Qwen3-VL has 258,400 tokens, within its context of 262,144, so no input is truncated. Dense sampling reaches the cap on 1,371 of the 2,697 Video-MME questions and on 704 of the 1,337 LongVideoBench questions, and \methodshort{} on 1,011 and 561. When \methodshort{} would keep more frames, we raise $\thresh$ for that video until the selection fits (\cref{sec:method:selection}). When dense sampling reaches the cap, it keeps 960 frames evenly spread over its 2\,FPS grid. The embedding-based selectors keep as many frames as \methodshort{}, while PySceneDetect and I-frames, which choose their own number of frames, thin their selection evenly to the cap on the 26 and 30 Video-MME videos where they exceed it.

\noindent\textbf{Choice of $\alpha$ and $\thresh$.}
We fix $\alpha = 0.5$ and $\thresh = 1$\,s for every LMM and benchmark, and vary them in \cref{tab:ablation,fig:threshold}.

\section{Benchmark Details}
\label{sec:app:datasets}

We use the versions of the benchmarks in VLMEvalKit. The short benchmarks have 12,061 questions on 6,572 videos, and the long ones 4,034 questions on 1,652 videos.

\noindent\textbf{TempCompass}~\cite{liu2024tempcompass} separates five temporal aspects, \ie, action, speed, direction, attribute change, and event order. It pairs videos with the same static content but different dynamics, so no question can be answered from a single frame. We use its multiple-choice, yes/no, and caption-matching formats, 5,536 questions on 410 videos, and exclude caption generation, which needs an LLM judge.

\noindent\textbf{MVBench}~\cite{li2024mvbench} turns image tasks into temporal ones, \eg, position into moving direction, and covers 20 tasks, from action counting and object shuffling to episodic reasoning and counterfactual inference. It has 4,000 multiple-choice questions on 3,871 videos.

\noindent\textbf{TVBench}~\cite{cores2025tvbench} is built to reduce what spatial cues and language priors can answer. Its wrong options are plausible events from the same video. It has 2,525 multiple-choice questions on 2,291 videos, across 10 tasks such as action sequence, moving direction, and object count.

\noindent\textbf{Video-MME}~\cite{fu2025video} has 2,700 multiple-choice questions on 900 videos from six domains, split equally into short (under 2 minutes), medium (4 to 15 minutes), and long (30 to 60 minutes) videos. We exclude the three questions on one long video (\texttt{tslKtm6Le1s}), since the video reader of VLMEvalKit fails on a late frame of it that MaxInfo selects, and report the remaining 2,697 questions for every method. On all 2,700 questions, dense sampling scores 71.33 and \methodshort{} 71.22, so the exclusion does not change their comparison. We do not use the subtitles.

\noindent\textbf{LongVideoBench}~\cite{wu2024longvideobench} asks questions that refer to a moment of the video, each testing either perception or the relation between moments. We use its 1,337 validation questions on 753 videos, 625 on perception and 712 on relations, in four duration groups, 8 to 15\,s (189 questions), 15 to 60\,s (172), 3 to 10 minutes (412), and 15 to 60 minutes (564). We do not use the subtitles.

\section{Evaluation Protocol}
\label{sec:app:metrics}
\noindent\textbf{Significance.}
To compare two methods on the same questions, we use the exact McNemar test, \ie, a two-sided binomial test on the questions that exactly one of the two methods answers correctly. To compare \methodshort{} over the twelve pairs of LMM and benchmark, with random frames (\cref{sec:exp:comparison}) and with the grid frames of highest novelty (\cref{sec:exp:ablation}), we use a one-sided paired $t$-test over the pairs. When we take each question as the unit instead (\cref{sec:app:tables}), we use a two-sided paired $t$-test over the questions.

\noindent\textbf{LMM time.}
We time the LMM on every question of each benchmark. The LMM generates a single token, so that the length of its answer does not affect the time. The time of a query is the wall-clock time of this call, synchronized with the GPU, and covers the vision encoder on the kept frames and the prefill, \ie, the time to the first token after preprocessing. Each run uses batch size 1 and bfloat16, and its first query is an untimed warm-up. Dense sampling and the methods it is compared with run back to back in the same job, on the same GPU. When a job cannot measure all methods within the time limit of the cluster, the remaining ones run in a second job that also measures dense sampling, and their LMM times are rescaled by the ratio between the dense-sampling times of the two jobs.

\noindent\textbf{Selection time and latency.}
Selection does not depend on the question, so it runs once per video, and each question on the video takes an equal share of its time. The latency of a query, as reported in our tables, is its LMM time plus this share, and we report the median over the questions. On the long-video benchmarks, we report the mean instead, since their questions fall into two groups, on short videos and on videos long enough to reach the 960-frame cap, and the median falls in the gap between them. For \methodshort{}, the selection time includes the signal export, measured as the extra decoding time it causes (\cref{sec:app:efficiency_breakdown}), as well as reading the signals and running the rule. The embedding-based selectors are timed as described in \cref{sec:app:baselines}. Decoding is needed by every method and is not counted. The only exception is PySceneDetect, which decodes the video with its own decoder and cannot reuse the decoding of the pipeline, so its selection time includes this decoding.

\noindent\textbf{Vision encoder time.}
We also time the vision encoder inside each timed call, with GPU-synchronized hooks, to obtain the latency of DPSelect with token reuse (\cref{sec:app:baselines}).

\noindent\textbf{Answer extraction.}
Qwen3.5 and Gemma~4 often explain their reasoning before answering, and the explanation mentions several options, which the exact matching of VLMEvalKit counts as a wrong answer even when the final answer is right. We therefore score every LMM and method on the final stated answer, with the same deterministic rule. It takes the text after the last occurrence of ``best option'', ``final answer'', ``the answer is'', ``correct answer is'', ``correct option is'', or ``answer:'', otherwise the whole prediction if it has at most 60 characters, otherwise its last non-empty line. The extracted answer is then scored by the unchanged evaluation of each benchmark.

\section{Baseline Details}
\label{sec:app:baselines}

MaxInfo, KTV, and DPSelect are re-implemented from their official code, and PySceneDetect runs through its own library. The selections used for accuracy follow that code, including its preprocessing. We time the selection of the embedding-based selectors separately, with an implementation optimized for speed. The decoded frames are moved to the GPU once, resized and normalized there, and encoded in batches in half precision. The selection algorithms themselves, \eg, the maximum-volume search of MaxInfo and the $k$-means of KTV, run as in the official code, on the CPU.

\noindent\textbf{Candidate frames.}
MaxInfo, DPSelect, and I-frames pick among the same candidates as \methodshort{}, the frames of the 2\,FPS grid used in our experiments. The code of MaxInfo samples at 1\,FPS, from which it cannot keep as many frames as \methodshort{} on most videos, so we use the 2\,FPS grid at the same number of frames. KTV takes every frame of the video, or 5,400 evenly spaced frames for longer videos, as in its code, and PySceneDetect analyzes every frame. These two can therefore keep frames off the grid.

\noindent\textbf{MaxInfo}~\cite{li2026maxinfo} embeds each candidate with CLIP ViT-L/14 at 336 pixels, takes the first eight singular vectors of the embeddings, and keeps the frames found by the rectangular maximum-volume algorithm, with the tolerance of its code (0.1). At its own setting, it samples the candidates at 1\,FPS and chooses how many frames to keep. On the short benchmarks it keeps all its candidates, so it is equivalent to uniform sampling at 1\,FPS.

\noindent\textbf{KTV}~\cite{song2026ktv} selects keyframes in a first stage that does not depend on the question, which is the stage we use. It embeds the candidates with DINOv2-L, clusters the embeddings with $k$-means into as many clusters as frames to keep, and keeps the frame closest to each centroid. When two centroids share their closest frame, its code falls back to evenly spaced frames, which happens for one of the 6,572 short-benchmark videos, 5 of the 900 Video-MME videos, and 4 of the 753 LongVideoBench videos. At its own setting it keeps 6 frames, as in its paper.

\noindent\textbf{DPSelect}~\cite{wang2024retake} encodes each candidate with the vision encoder of the evaluated LMM, with the same settings as in our evaluation, and measures the change between consecutive frames as the cosine distance between their visual tokens, averaged over the spatial positions. It keeps the local maxima of this change first, then the largest remaining changes. It has no frame count of its own, so it appears only at the same number of frames. In the official code, the selection runs inside the LMM, which reuses the visual tokens of the kept frames. We do the same in the latency, by subtracting the time the LMM spends encoding the kept frames, measured in the same call (\cref{sec:app:metrics}).

\noindent\textbf{PySceneDetect}~\cite{castellano2026pyscenedetect} detects shot boundaries with its content detector at the default threshold, decoding the video itself, and keeps the middle frame of each shot.

\noindent\textbf{I-frames} keeps the keyframes of the stream, moved to the nearest grid frame. The decoder reports the frame types, so selection costs nothing beyond decoding.

\noindent\textbf{Same number of frames.}
In the main comparison, MaxInfo, KTV, and DPSelect keep, for each video, as many frames as \methodshort{}, while PySceneDetect and I-frames run at their own setting. \cref{tab:main_comparison} also reports these two at the number of frames of \methodshort{}, for which we adapt them as follows. PySceneDetect gives each shot one frame and shares the rest among the shots in proportion to their length, spread evenly within each shot. It finds a single shot in most short clips (81 to 85\%), so on them it keeps evenly spaced frames. Some long videos have more shots than frames to keep, and it then keeps the middle frame of the longest shots. I-frames starts from the keyframes. When they are fewer than the frames to keep, it repeatedly adds the grid frame in the middle of the largest gap. When they are more, it keeps an evenly spaced subset of them.

\noindent\textbf{Controls.}
\textit{Random} keeps, for each video, as many frames as \methodshort{}, drawn uniformly from the 2\,FPS grid, with five seeds on short and three on long videos. \textit{Evenly spaced} keeps the same number of frames evenly spread over the grid, including its first and last frame.

\section{Gradual Change Missed by the Codec}
\label{sec:app:selection_strategies}

\noindent\textbf{Content drift in segments the codec calls static.}
We test whether the content still changes where the codec signals register little change. On the 2\,FPS grid, we take the one-second window that starts at each grid frame, 193,771 windows on 6,548 short-benchmark videos (the other 24 are shorter than 2\,s). As defined in \cref{sec:method:selection}, a window with the average novelty of its video has 1\,s of codec evidence. We measure its content change as the cosine distance between the SigLIP~2 embeddings of its two end frames, relative to a typical second, \ie, the median window of the same video. Windows with less than a quarter of the average codec evidence, \ie, below 0.25\,s, are 6.2\% of all windows. Their content still changes by a median of 43\% of a typical second. On the other hand, content change grows with codec evidence. When we split the windows into five equal groups by codec evidence, the median change rises from 58\% of a typical second in the group with the least evidence to 159\% in the group with the most. The codec signals thus track how much the content changes, but miss part of the slow change, which is why the selection rule also counts the elapsed time.

\begin{table}[t]
\centering
\caption{\textbf{Rule without the elapsed time.} Accuracy (\%) of Qwen3-VL-8B. Kept: share of the 2\,FPS grid. $^*$: significantly different (McNemar, $p < 0.05$) from the default (first two rows) or from $\alpha = 0.5$ at $\thresh = 0.5$\,s (other rows).}
\label{tab:ablation_notime}
\resizebox{\linewidth}{!}{
\begin{tabular}{@{}l ccc cc@{}}
\toprule
& \textbf{TC} & \textbf{MVB} & \textbf{TVB} & \textbf{Avg} & \textbf{Kept} \\
\midrule
\rowcolor{oursbg} Default ($\alpha = 0.5$, $\thresh = 1$\,s) & 75.9 & 66.5 & 59.3 & 67.2 & 64\% \\
No elapsed time & 75.3 & 64.7$^*$ & 57.1$^*$ & 65.7 & 36\% \\
\midrule
\multicolumn{6}{@{}l}{\textit{No elapsed time, $\thresh = 0.5$\,s}} \\
$\alpha = 0.5$ & 75.3 & 65.9 & 58.8 & 66.7 & 58\% \\
\multicolumn{6}{@{}l}{\textit{No elapsed time, same number of frames}} \\
$\alpha = 0$ ($E_I$ only) & 75.0 & 65.4 & 58.9 & 66.4 & 58\% \\
$\alpha = 0.2$ & 75.3 & 66.1 & 59.7$^*$ & 67.0 & 58\% \\
$\alpha = 0.8$ & 75.5 & 65.8 & 59.2 & 66.9 & 58\% \\
$\alpha = 1$ ($E_R$ only) & 75.5 & 65.5 & 58.5 & 66.5 & 58\% \\
\bottomrule
\end{tabular}}
\end{table}

\noindent\textbf{Rule without the elapsed time.}
Without the elapsed time, a frame is kept only when the codec evidence alone reaches $\thresh$, so segments the codec registers as static receive no frames at all (\cref{tab:ablation_notime}). At the same $\thresh = 1$\,s, the rule then keeps 36\% of the grid and loses 1.5 points on average, with significant drops on MVBench ($-1.8$) and TVBench ($-2.2$). Halving $\thresh$ brings it back to 58\% of the grid, still about half a point below the default, and significantly so on TempCompass. Part of these drops comes from keeping fewer frames. To compare the two signals when the novelty decides every kept frame, we also set $\thresh$ for each $\alpha$ so all variants keep 58\% of the grid. The single signals are then the least accurate variants (66.4 for $E_I$ and 66.5 for $E_R$), and every combination is above both (66.7 to 67.0), by a small margin of 0.2 to 0.6 points. We keep $\alpha = 0.5$, which weighs the two signals equally, although $\alpha = 0.2$ is slightly better here.

\section{Efficiency Breakdown}
\label{sec:app:efficiency_breakdown}

\begin{figure}[t]
    \centering
    \includegraphics[width=\linewidth]{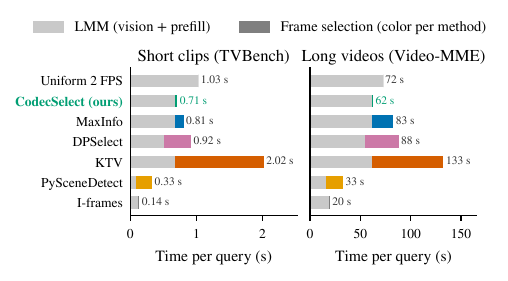}
    \caption{\textbf{Where the time goes.} Time per query with Qwen3-VL-8B on TVBench and Video-MME, in the settings of \cref{tab:main,tab:videomme}. Gray: the LMM (vision encoder and prefill). Color: frame selection, including the signal export of \methodshort{} and, for DPSelect, the encoding of the kept frames, whose visual tokens it reuses. Shared decoding is not shown.}
    \label{fig:breakdown}
\end{figure}

\noindent\textbf{Cost of the codec signals.}
The decoder computes the codec signals anyway, but exporting them takes time. We measure it on all 8,225 test videos by decoding each with FFmpeg on 8 threads of an Intel Xeon Platinum 8358 CPU, with and without the export. The export adds a median of 8\% to decoding, 4 to 12\,ms per short clip and 0.14 to 0.27\,s per long video. Reading the signals and running the rule take a median of 0.015 to 0.14\,s per video. Every latency we report includes both. As a share of the total time per query, decoding included, the export is 0.5 to 1.4\% (\cref{tab:decoding_cost}). The selection cost of \methodshort{} is thus 2 to 10 times lower than that of the embedding-based selectors on clips shorter than 10\,s, and 31 to 84 times lower on videos longer than 10 minutes (\cref{fig:scaling}), since they encode every candidate frame while \methodshort{} reads one number per frame. 
\cref{fig:breakdown} splits the time per query into the LMM and frame selection.

\begin{table}[t]
\centering\small\setlength{\tabcolsep}{2pt}
\caption{\textbf{Cost of the codec signals.} Medians per video. Rule: reading the signals and selecting. Export / total: share of the total time per query (decoding, selection, and LMM), range over the four LMMs. $^\ast$Qwen3-VL-8B only, with mean times as in \cref{tab:videomme}.}
\label{tab:decoding_cost}
\resizebox{\linewidth}{!}{
\begin{tabular}{@{}l r ccc c c@{}}
\toprule
\textbf{Benchmark} & \textbf{Videos} & \textbf{Decoding} & \textbf{Export} & \shortstack{\textbf{Export}\\\textbf{/ dec.}} & \textbf{Rule} & \shortstack{\textbf{Export}\\\textbf{/ total}} \\
\midrule
TempCompass & 410 & 0.061\,s & 0.005\,s & 8.5\% & 0.016\,s & 0.7--1.4\% \\
MVBench & 3,871 & 0.055\,s & 0.004\,s & 8.5\% & 0.015\,s & 0.5--1.3\% \\
TVBench & 2,291 & 0.173\,s & 0.012\,s & 7.4\% & 0.017\,s & 0.6--1.3\% \\
Video-MME & 900 & 4.26\,s & 0.27\,s & 6.7\% & 0.14\,s & 0.9\%$^\ast$ \\
LongVideoBench & 753 & 1.53\,s & 0.14\,s & 7.0\% & 0.05\,s & 0.8\%$^\ast$ \\
\bottomrule
\end{tabular}}
\end{table}

\section{Detailed Results}
\label{sec:app:tables}

This section gives the results of \cref{sec:exp} in full, per LMM, at other frame rates, on long videos, and for other $\alpha$ and $\thresh$.

\subsection{Per-LMM and Per-Benchmark Results}

\begin{figure*}[t]
    \centering
    \includegraphics[width=\textwidth]{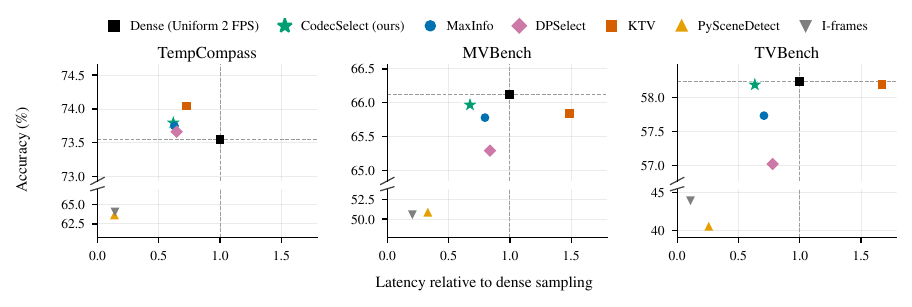}
    \caption{\textbf{Accuracy vs latency, per benchmark}, averaged over the four LMMs, as in \cref{fig:main_tradeoff}. Latency relative to dense sampling. PySceneDetect and I-frames, far less accurate, lie below a break in the accuracy axis. Above the break, all panels share one scale.}
    \label{fig:main_tradeoff_bench}
\end{figure*}

\cref{tab:main_comparison} gives the comparison of \cref{tab:main} with the latency of each LMM and every baseline, the fast selectors included, at the same number of frames as \methodshort{}. \cref{fig:main_tradeoff_bench} shows the trade-off of \cref{fig:main_tradeoff} separately for each benchmark. \cref{tab:native_settings} gives the baselines at their own setting, with different numbers of frames, hence different points of the trade-off between frames and accuracy rather than a comparison at the same cost. On the short benchmarks, MaxInfo keeps all its 1\,FPS candidates and thus gives exactly the accuracy of uniform sampling at 1\,FPS (\cref{tab:uniform_rates}), at the added cost of its selection. KTV is below dense sampling (\cref{tab:main_comparison}) with three of the four LMMs, and PySceneDetect and I-frames are 8.5 to 16.5 points below it.

\subsection{Controls}

\begin{table}[t]
\centering
\caption{\textbf{Controls per LMM.} Accuracy (\%) of \methodshort{} and of the two controls of \cref{tab:main}. Random: mean $\pm$ standard deviation over five seeds. $^\dagger$: significantly different from \methodshort{} in all five seeds (McNemar, $p < 0.05$).}
\label{tab:controls_per_lmm}
\resizebox{\linewidth}{!}{
\begin{tabular}{@{}ll ccc@{}}
\toprule
\textbf{LMM} & \textbf{Benchmark} & \textbf{\methodshort{}} & \textbf{Random} & \textbf{Evenly} \\
\midrule
Qwen3-VL-8B & TempCompass & 75.9 & 75.5\,{\scriptsize$\pm$0.3} & 75.8 \\
Qwen3-VL-8B & MVBench & 66.5 & 65.5\,{\scriptsize$\pm$0.3} & 66.5 \\
Qwen3-VL-8B & TVBench & 59.3 & 58.2\,{\scriptsize$\pm$0.3} & 58.9 \\
Qwen3.5-9B & TempCompass & 74.0 & 73.8\,{\scriptsize$\pm$0.3} & 73.6 \\
Qwen3.5-9B & MVBench & 67.0 & 66.5\,{\scriptsize$\pm$0.3} & 67.0 \\
Qwen3.5-9B & TVBench & 60.8 & 59.9\,{\scriptsize$\pm$0.2} & 60.6 \\
InternVL3.5-8B & TempCompass & 71.2 & 70.8\,{\scriptsize$\pm$0.3} & 71.3 \\
InternVL3.5-8B & MVBench & 72.6 & 71.3\,{\scriptsize$\pm$0.2}$^\dagger$ & 72.4 \\
InternVL3.5-8B & TVBench & 61.6 & 61.0\,{\scriptsize$\pm$0.3} & 62.2 \\
Gemma-4-12B & TempCompass & 74.0 & 73.5\,{\scriptsize$\pm$0.3} & 74.2 \\
Gemma-4-12B & MVBench & 57.8 & 56.8\,{\scriptsize$\pm$0.2} & 57.3 \\
Gemma-4-12B & TVBench & 51.0 & 51.0\,{\scriptsize$\pm$0.3} & 52.2 \\
\bottomrule
\end{tabular}}
\end{table}

\begin{table*}[t]
\centering\small\setlength{\tabcolsep}{3.2pt}
\caption{\textbf{Uniform sampling at lower rates.} Accuracy (\%). Uniform 1.28\,FPS keeps as many frames in total as \methodshort{}. Frames: relative to dense sampling. $^\dagger$: significantly different from dense sampling (McNemar, $p < 0.05$).}
\label{tab:uniform_rates}
\begin{tabular}{@{}l ccc ccc ccc ccc c@{}}
\toprule
 & \multicolumn{3}{c}{\textbf{Qwen3-VL}} & \multicolumn{3}{c}{\textbf{Qwen3.5}} & \multicolumn{3}{c}{\textbf{InternVL3.5}} & \multicolumn{3}{c}{\textbf{Gemma~4}} & \\
\cmidrule(lr){2-4} \cmidrule(lr){5-7} \cmidrule(lr){8-10} \cmidrule(lr){11-13}
\textbf{Method} & TC & MVB & TVB & TC & MVB & TVB & TC & MVB & TVB & TC & MVB & TVB & \textbf{Frames} \\
\midrule
\rowcolor{baselinebg} Dense (Uniform 2\,FPS) & 75.9 & 67.1 & 59.8 & 73.5 & 67.7 & 61.5 & 71.7 & 72.9 & 62.1 & 73.1 & 56.9 & 49.5 & 100\% \\
\rowcolor{oursbg} \textbf{\methodshort{} (ours)} & 75.9 & 66.5 & 59.3 & 74.0 & 67.0 & 60.8 & 71.2 & 72.6 & 61.6 & 74.0$^\dagger$ & 57.8 & 51.0$^\dagger$ & 64\% \\
Uniform 1.28\,FPS & 75.8 & 66.5 & 59.0 & 73.5 & 67.2 & 60.5 & 71.2 & 72.4 & 61.5 & 74.1$^\dagger$ & 57.5 & 51.2$^\dagger$ & 65\% \\
Uniform 1\,FPS & 75.6 & 66.0$^\dagger$ & 59.4 & 74.1 & 66.4$^\dagger$ & 60.8 & 70.8$^\dagger$ & 72.4 & 60.6$^\dagger$ & 74.3$^\dagger$ & 57.9 & 52.3$^\dagger$ & 51\% \\
\bottomrule
\end{tabular}
\end{table*}

\noindent\textbf{Random frames.}
\cref{tab:controls_per_lmm} gives the two controls of \cref{tab:main} for each LMM. \methodshort{} is more accurate than random frames at the same number of frames in all twelve pairs of LMM and benchmark (\cref{sec:exp:comparison}). Since the four LMMs answer the same questions, the twelve pairs are not fully independent, so we also take each question as the unit, with its difference averaged over the LMMs and seeds. \methodshort{} is then ahead by 0.64 points over 12,061 questions ($p < 10^{-9}$), and significantly on each benchmark.

\noindent\textbf{Lower fixed rates.}
\methodshort{} does not aim to beat uniform sampling at a lower rate (\cref{tab:uniform_rates}), but to keep the accuracy of the dense rate it starts from. Uniform sampling matches dense sampling only at 1.28\,FPS, a rate we set to the total number of frames of \methodshort{}, which is not known in advance. At 1\,FPS, it is significantly below dense sampling in four of the twelve pairs of LMM and benchmark, while \methodshort{} is never significantly below it.

\subsection{Long Videos}\label{sec:app:long_videos}

\begin{table}[t]
\centering\small\setlength{\tabcolsep}{3.5pt}
\caption{\textbf{Controls on long videos.} Accuracy (\%) of Qwen3-VL-8B. Fr.: frames per question. Uniform, same total frames: 1.28\,FPS on Video-MME and 1.24\,FPS on LongVideoBench. $^\dagger$: significantly different from dense sampling (McNemar, $p < 0.05$).}
\label{tab:long_controls}
\begin{tabular}{@{}l cc cc@{}}
\toprule
 & \multicolumn{2}{c}{\textbf{Video-MME}} & \multicolumn{2}{c}{\textbf{LongVideoBench}} \\
\cmidrule(lr){2-3} \cmidrule(lr){4-5}
\textbf{Method} & Acc. & Fr. & Acc. & Fr. \\
\midrule
\rowcolor{baselinebg} Dense (Uniform 2\,FPS) & 71.3 & 654 & 65.9 & 647 \\
\rowcolor{oursbg} \textbf{\methodshort{} (ours)} & 71.2 & 569 & 67.4$^\dagger$ & 563 \\
Uniform, same total frames & 71.0 & 570 & 65.7 & 563 \\
Uniform 1\,FPS & 71.3 & 518 & 64.9 & 530 \\
Evenly spaced & 71.3 & 569 & 66.5 & 563 \\
Random (3 seeds) & 71.0 & 569 & 65.6 & 563 \\
\bottomrule
\end{tabular}
\end{table}

\cref{tab:long_controls} adds uniform sampling at 1\,FPS and the two controls at the per-video number of frames of \methodshort{}. On Video-MME, none of them differs significantly from dense sampling or from \methodshort{}, with random frames at 71.1, 70.8, and 71.2 over the three seeds. On LongVideoBench, \methodshort{} is significantly more accurate than dense sampling and than uniform sampling at 1.24 and 1\,FPS, and more accurate than random frames in all three seeds (65.5, 66.2, and 65.3), significantly in two. Evenly spaced frames lie in between, not significantly different from \methodshort{}.

\noindent\textbf{Original encoding.}
On the 232 Temporal Perception and Temporal Reasoning questions of Video-MME, with the original videos instead of re-encoded ones, \methodshort{} matches dense sampling with Qwen3-VL-8B (64.7 for both) while using 90\% of its frames, more than on short clips, as many long videos reach the 960-frame cap.

\subsection{Denser Grid}

\begin{table}[t]
\centering\small\setlength{\tabcolsep}{3.5pt}
\caption{\textbf{\methodshort{} on a 4\,FPS grid}, with $\thresh = 1$\,s. Accuracy (\%). \methodshort{} keeps 38\% of the grid. $^\dagger$: significantly different from dense 4\,FPS sampling (McNemar, $p < 0.05$).}
\label{tab:fourfps}
\begin{tabular}{@{}ll cccc@{}}
\toprule
\textbf{LMM} & \textbf{Method} & \textbf{TC} & \textbf{MVB} & \textbf{TVB} & \textbf{Avg} \\
\midrule
Qwen3-VL-8B & Dense 4\,FPS & 76.2 & 67.9 & 59.6 & 67.9 \\
 \rowcolor{oursbg} & \textbf{\methodshort{}} & 76.4 & 67.1 & 59.4 & 67.6 \\
\addlinespace[2pt]
Qwen3.5-9B & Dense 4\,FPS & 72.7 & 68.1 & 61.7 & 67.5 \\
 \rowcolor{oursbg} & \textbf{\methodshort{}} & 73.3 & 67.3 & 61.8 & 67.5 \\
\addlinespace[2pt]
InternVL3.5-8B & Dense 4\,FPS & 72.5 & 73.0 & 60.1 & 68.5 \\
 \rowcolor{oursbg} & \textbf{\methodshort{}} & 71.4$^\dagger$ & 73.0 & 61.4 & 68.6 \\
\bottomrule
\end{tabular}
\end{table}

\cref{tab:fourfps} details the 4\,FPS result of \cref{sec:exp:comparison}. With the same $\thresh$, \methodshort{} keeps 38\% of the 4\,FPS grid, about three quarters of the frames of dense 2\,FPS sampling, and matches dense 4\,FPS sampling on average (67.9 against 68.0), with one significant drop, InternVL3.5 on TempCompass. Gemma~4 is not included, since dense 4\,FPS sampling with it does not fit in the memory of our GPUs.

\subsection{Weight of the Two Signals}

\begin{table}[t]
\centering
\caption{\textbf{Weight of the two signals.} Accuracy (\%) averaged over the four LMMs. $\alpha$: weight of $E_R$ in the novelty. Kept: share of the 2\,FPS grid. Drops: pairs of LMM and benchmark, out of 12, significantly below dense sampling (McNemar, $p < 0.05$).}
\label{tab:ablation}
\resizebox{\linewidth}{!}{
\begin{tabular}{@{}l ccc ccc@{}}
\toprule
& \textbf{TC} & \textbf{MVB} & \textbf{TVB} & \textbf{Avg} & \textbf{Kept} & \textbf{Drops} \\
\midrule
$\alpha = 0$ ($E_I$ only) & 73.9 & 65.7 & 58.2 & 65.95 & 61\% & 2 \\
$\alpha = 0.2$ & 73.8 & 66.0 & 58.1 & 65.97 & 62\% & 2 \\
\rowcolor{oursbg} $\alpha = 0.5$ (default) & 73.8 & 66.0 & 58.2 & 65.98 & 64\% & \textbf{0} \\
$\alpha = 0.8$ & 73.7 & 66.0 & 58.2 & 65.95 & 65\% & 1 \\
$\alpha = 1$ ($E_R$ only) & 73.6 & 65.9 & 58.6 & 66.04 & 65\% & 2 \\
\bottomrule
\end{tabular}}
\end{table}

Accuracy is flat in $\alpha$ on average (\cref{tab:ablation}), from 65.95 to 66.04 over the four LMMs and three benchmarks. The difference lies in reliability. With $\alpha = 0.5$, \methodshort{} is never significantly below dense sampling in any of the twelve pairs of LMM and benchmark, while each signal alone falls significantly below it in two. Combining the two signals avoids these drops.

\subsection{Threshold}

\begin{figure}[t]
    \centering
    \includegraphics[width=\linewidth]{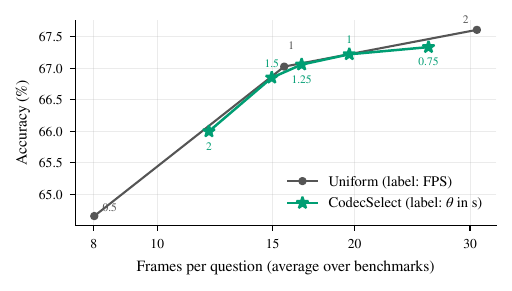}
    \caption{\textbf{Threshold.} Accuracy of Qwen3-VL-8B against frames per question, averaged over the three short benchmarks, for uniform sampling and for \methodshort{} with $\thresh$ from 0.75 to 2\,s.}
    \label{fig:threshold}
\end{figure}

In \cref{fig:threshold}, we vary $\thresh$ from 0.75 to 2\,s, with uniform sampling at 0.5 to 2\,FPS for reference. Accuracy changes smoothly with $\thresh$, from 67.3 at 0.75\,s to 67.2 at the default of 1\,s and 66.0 at 2\,s, as the number of frames per question falls from 26 to 12. Up to 1.5\,s, it stays within 0.4 points of the default.

\subsection{Rule against Top-$k$ Selection}

\begin{table}[t]
\centering\small\setlength{\tabcolsep}{4pt}
\caption{\textbf{Rule against top-$k$ selection.} Accuracy (\%) averaged over the four LMMs, with as many frames per video as \methodshort{}. Coverage: loss of \cref{fig:coverage} ($\times 1000$).}
\label{tab:topk}
\resizebox{\linewidth}{!}{
\begin{tabular}{@{}l cccc c@{}}
\toprule
\textbf{Method} & \textbf{TC} & \textbf{MVB} & \textbf{TVB} & \textbf{Avg} & \textbf{Coverage} \\
\midrule
\rowcolor{baselinebg} Dense (Uniform 2\,FPS) & 73.6 & 66.1 & 58.2 & 65.97 & -- \\
\rowcolor{oursbg} \textbf{\methodshort{} (ours)} & 73.8 & 66.0 & 58.2 & 65.98 & 16.31 \\
Top-$k$, point & 73.6 & 65.5 & 57.6 & 65.55 & 18.92 \\
Top-$k$, interval & 73.3 & 65.3 & 57.4 & 65.35 & 19.53 \\
\bottomrule
\end{tabular}}
\end{table}

Instead of the rule, one can keep the grid frames with the highest novelty, as many per video as \methodshort{}. We score a grid frame either by its own novelty $\score_t$ (point) or by the novelty summed since the previous grid frame, as in the codec evidence (interval). Both are less accurate than \methodshort{} (\cref{tab:topk}), by 0.44 points for point, lower in 10 of the 12 pairs, and by 0.64 points for interval, lower in 11 of the 12 pairs and significantly so in two. Both also cover the video worse, with a 16\% and a 20\% higher coverage loss, since the most novel frames cluster around cuts. The longest gap between kept frames exceeds 2\,s in 34\% and 55\% of the videos, and in none for \methodshort{}.

\section{Visualizations}
\label{sec:app:visualizations}

\begin{figure*}[t]
    \centering
    \includegraphics[width=\textwidth]{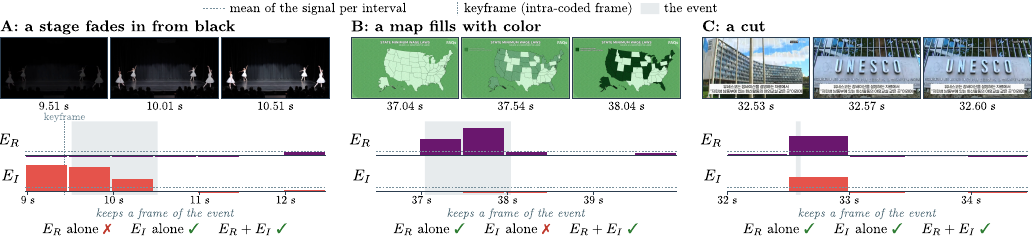}
    \caption{\textbf{Each signal catches an event the other one misses.} $E_R$ and $E_I$ per interval of the 2\,FPS grid, relative to their mean (dotted line), and whether each signal alone or their combination, without the elapsed time (\cref{tab:buildup_count}), keeps a frame of the event. \textbf{(A)} A fade-in is coded intra with little residual. \textbf{(B)} A map filling with color is predicted with a large residual. \textbf{(C)} A cut raises both.}
    \label{fig:qualitative}
\end{figure*}

\cref{fig:qualitative} shows three events, two that only one codec signal registers and a cut that both register.

The project page \url{https://lucazanella.github.io/codecselect/} shows the frames that \methodshort{} keeps on ten videos from the five benchmarks, with the answers of Qwen3-VL-8B. For each video, it shows all the frames of the 2\,FPS grid and, as in \cref{fig:method_pipeline}, the evidence since the last kept frame. \methodshort{} keeps frames densely while the codec registers change and never leaves a gap longer than $\thresh$ (here 1\,s). The last two examples are failure cases, in which only \methodshort{} answers wrongly although its kept frames show the answer.

\section{Limitations}
\label{sec:app:limitations}

\methodshort{} normalizes its signals over the whole video, so it selects once the video is available, as in indexing and offline pipelines, and running estimates of these statistics could extend it to live streams. The signals also require a decoder that exposes them, which hardware decoders (\eg, NVDEC) currently do not.
\section{Broader Impact}
\label{sec:app:broader_impact}

\methodshort{} lowers the cost of video understanding with LMMs, as it removes redundant frames before any neural network processes them, without encoding any frame itself. Less computation per query makes long videos and large collections cheaper to process, \eg, for indexing, and reduces the energy spent for the same accuracy. Because the selection needs no neural model, it can also run where the video is decoded, \eg, on a device, before frames are sent to a remote LMM. The same reduction in cost applies to any video analysis, including the monitoring of people, so systems using it need the usual safeguards on consent, privacy, and appropriate use. \methodshort{} does not change what an LMM can infer from the frames it receives, and it inherits the biases of the LMMs and benchmarks it is used with.

\begin{table}[t]
\centering\small\setlength{\tabcolsep}{4pt}
\caption{\textbf{Assets and licenses.} Licenses as stated by their owners.}
\label{tab:assets}
\resizebox{\linewidth}{!}{
\begin{tabular}{@{}ll@{}}
\toprule
\textbf{Asset} & \textbf{License} \\
\midrule
\multicolumn{2}{@{}l}{\textit{Benchmarks}} \\
TempCompass~\cite{liu2024tempcompass} & CC BY-NC 4.0 \\
MVBench~\cite{li2024mvbench} & MIT \\
TVBench~\cite{cores2025tvbench} & CC BY 4.0 \\
Video-MME~\cite{fu2025video} & Academic research only \\
LongVideoBench~\cite{wu2024longvideobench} & CC BY-NC-SA 4.0 \\
NTU RGB+D clips (in MVBench, TVBench) & ROSE Lab terms of use \\
\addlinespace[2pt]
\multicolumn{2}{@{}l}{\textit{Models}} \\
Qwen3-VL-8B-Instruct~\cite{bai2025qwen3} & Apache 2.0 \\
Qwen3.5-9B~\cite{qwen35blog} & Apache 2.0 \\
InternVL3.5-8B~\cite{wang2025internvl3} & Apache 2.0 \\
Gemma-4-12B~\cite{team2026gemma} & Apache 2.0 \\
CLIP ViT-L/14 at 336 pixels~\cite{radford2021learning} & MIT \\
DINOv2-L~\cite{oquab2024dinov2} & Apache 2.0 \\
SigLIP~2 So400m/14 at 384 pixels~\cite{tschannen2025siglip} & Apache 2.0 \\
\addlinespace[2pt]
\multicolumn{2}{@{}l}{\textit{Code of the baselines}} \\
MaxInfo~\cite{li2026maxinfo} & Not stated \\
KTV~\cite{song2026ktv} & Not stated \\
ReTaKe (DPSelect)~\cite{wang2024retake} & MIT \\
\addlinespace[2pt]
\multicolumn{2}{@{}l}{\textit{Software}} \\
VLMEvalKit & Apache 2.0 \\
FFmpeg 9.0.2 with libx264 and libx265 & GPL 2.0 or later \\
PySceneDetect~\cite{castellano2026pyscenedetect} & BSD 3-Clause \\
Decord & Apache 2.0 \\
PyTorch & BSD-style \\
Transformers & Apache 2.0 \\
scikit-learn & BSD 3-Clause \\
maxvolpy & MIT \\
\bottomrule
\end{tabular}}
\end{table}

\section{Assets}
\label{sec:app:assets}

\cref{tab:assets} lists the benchmarks, models, and software we use, with their licenses. Models and benchmarks are taken from Hugging Face. The clips of NTU RGB+D in MVBench and TVBench are subject to the terms of use of the ROSE Lab.

\begin{table*}[p]
\centering%
\caption{\textbf{Comparison at the same number of frames, per LMM.} Every baseline keeps, for each video, as many frames as \methodshort{}. Frames: relative to dense sampling. Sel.: mean selection time per video. Latency: median time per query relative to dense sampling, averaged over the benchmarks. $^\dagger$: significantly different from \methodshort{} (McNemar, $p < 0.05$).}
\label{tab:main_comparison}
\begin{tabular}{@{}l cccc ccc@{}}
\toprule
\multirow{2}{*}{\textbf{Method}} & \multicolumn{4}{c}{\textbf{Accuracy (\%)}} & \multirow{2}{*}{\textbf{Frames}} & \multirow{2}{*}{\textbf{Sel. (s)}} & \multirow{2}{*}{\textbf{Latency}} \\
\cmidrule(lr){2-5}
 & \textbf{TC} & \textbf{MVB} & \textbf{TVB} & \textbf{Avg.} & & & \\
\midrule
\multicolumn{8}{@{}l}{\textit{\textbf{Qwen3-VL-8B}}} \\
\rowcolor{baselinebg} Dense (Uniform 2\,FPS) & 75.9 & 67.1 & 59.8 & 67.6 & 100\% & 0 & 1.00$\times$ \\
 MaxInfo~\cite{li2026maxinfo} & 75.5 & 66.4 & 58.4 & 66.8 & 64\% & 0.12 & 0.78$\times$ \\
 KTV~\cite{song2026ktv} & 76.0 & 66.7 & 59.0 & 67.3 & 64\% & 1.24 & 1.55$\times$ \\
 DPSelect~\cite{wang2024retake} & 75.4$^\dagger$ & 66.1 & 59.4 & 67.0 & 64\% & 0.40 & 0.85$\times$ \\
 PySceneDetect~\cite{castellano2026pyscenedetect} & 75.7 & 66.5 & 58.5 & 66.9 & 64\% & 0.29 & 0.83$\times$ \\
 I-frames~\cite{dong2026rethinking} & 75.4 & 65.5$^\dagger$ & 58.2 & 66.4 & 64\% & 0.02 & 0.69$\times$ \\
\rowcolor{oursbg} \textbf{\methodshort{} (ours)} & 75.9 & 66.5 & 59.3 & 67.2 & 64\% & 0.03 & 0.69$\times$ \\
\addlinespace
\multicolumn{8}{@{}l}{\textit{\textbf{Qwen3.5-9B}}} \\
\rowcolor{baselinebg} Dense (Uniform 2\,FPS) & 73.5 & 67.7 & 61.5 & 67.5 & 100\% & 0 & 1.00$\times$ \\
 MaxInfo~\cite{li2026maxinfo} & 74.0 & 67.2 & 60.0 & 67.0 & 64\% & 0.12 & 0.79$\times$ \\
 KTV~\cite{song2026ktv} & 74.3 & 66.8 & 60.6 & 67.2 & 64\% & 1.24 & 1.51$\times$ \\
 DPSelect~\cite{wang2024retake} & 73.9 & 67.0 & 59.6 & 66.8 & 64\% & 0.36 & 0.83$\times$ \\
 PySceneDetect~\cite{castellano2026pyscenedetect} & 74.2 & 66.8 & 60.9 & 67.3 & 64\% & 0.29 & 0.83$\times$ \\
 I-frames~\cite{dong2026rethinking} & 73.4 & 66.7 & 60.8 & 67.0 & 64\% & 0.02 & 0.68$\times$ \\
\rowcolor{oursbg} \textbf{\methodshort{} (ours)} & 74.0 & 67.0 & 60.8 & 67.3 & 64\% & 0.03 & 0.69$\times$ \\
\addlinespace
\multicolumn{8}{@{}l}{\textit{\textbf{InternVL3.5-8B}}} \\
\rowcolor{baselinebg} Dense (Uniform 2\,FPS) & 71.7 & 72.9 & 62.1 & 68.9 & 100\% & 0 & 1.00$\times$ \\
 MaxInfo~\cite{li2026maxinfo} & 71.3 & 72.4 & 60.8 & 68.2 & 64\% & 0.12 & 0.71$\times$ \\
 KTV~\cite{song2026ktv} & 71.8 & 71.9 & 61.8 & 68.5 & 64\% & 1.24 & 1.28$\times$ \\
 DPSelect~\cite{wang2024retake} & 71.3 & 71.4$^\dagger$ & 59.8$^\dagger$ & 67.5 & 64\% & 0.26 & 0.75$\times$ \\
 PySceneDetect~\cite{castellano2026pyscenedetect} & 71.1 & 72.3 & 61.5 & 68.3 & 64\% & 0.29 & 0.77$\times$ \\
 I-frames~\cite{dong2026rethinking} & 71.1 & 71.5$^\dagger$ & 61.3 & 68.0 & 64\% & 0.02 & 0.64$\times$ \\
\rowcolor{oursbg} \textbf{\methodshort{} (ours)} & 71.2 & 72.6 & 61.6 & 68.5 & 64\% & 0.03 & 0.65$\times$ \\
\addlinespace
\multicolumn{8}{@{}l}{\textit{\textbf{Gemma-4-12B}}} \\
\rowcolor{baselinebg} Dense (Uniform 2\,FPS) & 73.1$^\dagger$ & 56.9 & 49.5$^\dagger$ & 59.8 & 100\% & 0 & 1.00$\times$ \\
 MaxInfo~\cite{li2026maxinfo} & 74.2 & 57.2 & 51.8 & 61.1 & 64\% & 0.12 & 0.57$\times$ \\
 KTV~\cite{song2026ktv} & 74.0 & 58.1 & 51.4 & 61.1 & 64\% & 1.24 & 0.85$\times$ \\
 DPSelect~\cite{wang2024retake} & 74.1 & 56.7$^\dagger$ & 49.2$^\dagger$ & 60.0 & 64\% & 0.15 & 0.58$\times$ \\
 PySceneDetect~\cite{castellano2026pyscenedetect} & 74.4 & 56.8 & 51.6 & 61.0 & 64\% & 0.29 & 0.60$\times$ \\
 I-frames~\cite{dong2026rethinking} & 73.8 & 57.4 & 51.2 & 60.8 & 64\% & 0.02 & 0.54$\times$ \\
\rowcolor{oursbg} \textbf{\methodshort{} (ours)} & 74.0 & 57.8 & 51.0 & 61.0 & 64\% & 0.03 & 0.54$\times$ \\
\bottomrule
\end{tabular}%
\end{table*}

\begin{table*}[p]
\centering%
\caption{\textbf{Baselines at their own setting, per LMM}, as described in \cref{sec:app:baselines}. Frames: frames per question. Sel.: selection time per video. Latency: median time per query. All three are means over the benchmarks. DPSelect has no frame count of its own.}
\label{tab:native_settings}
\begin{tabular}{@{}l cccc ccc@{}}
\toprule
\multirow{2}{*}{\textbf{Method}} & \multicolumn{4}{c}{\textbf{Accuracy (\%)}} & \multirow{2}{*}{\textbf{Frames}} & \multirow{2}{*}{\textbf{Sel. (s)}} & \multirow{2}{*}{\textbf{Latency (s)}} \\
\cmidrule(lr){2-5}
 & \textbf{TC} & \textbf{MVB} & \textbf{TVB} & \textbf{Avg.} & & & \\
\midrule
\multicolumn{8}{@{}l}{\textit{\textbf{Qwen3-VL-8B}}} \\
 MaxInfo~\cite{li2026maxinfo} & 75.6 & 66.0 & 59.4 & 67.0 & 15.6 & 0.06 & 0.40 \\
 KTV~\cite{song2026ktv} & 76.3 & 64.5 & 54.7 & 65.1 & 6.0 & 1.24 & 0.69 \\
 PySceneDetect~\cite{castellano2026pyscenedetect} & 64.2 & 51.0 & 41.4 & 52.2 & 1.4 & 0.29 & 0.20 \\
 I-frames~\cite{dong2026rethinking} & 65.1 & 49.6 & 42.9 & 52.5 & 2.7 & 0.02 & 0.11 \\
\addlinespace
\multicolumn{8}{@{}l}{\textit{\textbf{Qwen3.5-9B}}} \\
 MaxInfo~\cite{li2026maxinfo} & 74.1 & 66.4 & 60.8 & 67.1 & 15.6 & 0.06 & 0.37 \\
 KTV~\cite{song2026ktv} & 75.1 & 65.6 & 54.2 & 65.0 & 6.0 & 1.24 & 0.67 \\
 PySceneDetect~\cite{castellano2026pyscenedetect} & 64.5 & 52.0 & 40.5 & 52.3 & 1.4 & 0.29 & 0.21 \\
 I-frames~\cite{dong2026rethinking} & 64.8 & 52.8 & 45.3 & 54.3 & 2.7 & 0.02 & 0.12 \\
\addlinespace
\multicolumn{8}{@{}l}{\textit{\textbf{InternVL3.5-8B}}} \\
 MaxInfo~\cite{li2026maxinfo} & 70.8 & 72.4 & 60.6 & 67.9 & 15.6 & 0.06 & 0.43 \\
 KTV~\cite{song2026ktv} & 71.0 & 69.9 & 58.8 & 66.5 & 6.0 & 1.24 & 0.67 \\
 PySceneDetect~\cite{castellano2026pyscenedetect} & 63.2 & 52.3 & 41.6 & 52.4 & 1.4 & 0.29 & 0.19 \\
 I-frames~\cite{dong2026rethinking} & 62.5 & 52.1 & 45.0 & 53.2 & 2.7 & 0.02 & 0.10 \\
\addlinespace
\multicolumn{8}{@{}l}{\textit{\textbf{Gemma-4-12B}}} \\
 MaxInfo~\cite{li2026maxinfo} & 74.3 & 57.9 & 52.3 & 61.5 & 15.6 & 0.06 & 0.83 \\
 KTV~\cite{song2026ktv} & 75.1 & 56.9 & 51.6 & 61.2 & 6.0 & 1.24 & 0.80 \\
 PySceneDetect~\cite{castellano2026pyscenedetect} & 62.7 & 48.2 & 38.9 & 49.9 & 1.4 & 0.29 & 0.22 \\
 I-frames~\cite{dong2026rethinking} & 63.7 & 47.7 & 42.5 & 51.3 & 2.7 & 0.02 & 0.15 \\
\bottomrule
\end{tabular}%
\end{table*}

\end{document}